\documentclass[10pt,twocolumn,letterpaper]{article}

\usepackage{cvpr}
\usepackage{times}
\usepackage{epsfig}
\usepackage{graphicx}
\usepackage{amsmath}
\usepackage{amssymb}
\usepackage{multirow}
\usepackage{listings}
\usepackage{wasysym}
\usepackage[utf8]{inputenc}
\usepackage{color}
\usepackage{subfig}
\usepackage{soul}
\usepackage{mathtools}
\usepackage{algorithm}
\usepackage{algpseudocode}

\newcommand{\norm}[1]{\left\lVert#1\right\rVert}

\definecolor{mycyan}{RGB}{0,255,255}
\definecolor{mymagenta}{RGB}{255,0,255}

\usepackage[pagebackref=true,breaklinks=true,letterpaper=true,colorlinks,bookmarks=false]{hyperref}

\cvprfinalcopy 

\def\cvprPaperID{2231} 

\begin{document}

\title{Learning to Extract Semantic Structure from Documents\\Using Multimodal Fully Convolutional Neural Networks\vspace{-1em}}

\author{
Xiao Yang$^{1}$, Ersin Yumer$^{2}$, Paul Asente$^{2}$, Mike Kraley$^{3}$, Daniel Kifer$^{1}$, C. Lee Giles$^{1}$\\
$^{1}$The Pennsylvania State University $\quad$ $^{2}$Adobe Research $\quad$ $^{3}$Adobe Document Cloud\\
\small{xuy111@psu.edu \ \{yumer, asente, mkraley\}@adobe.com \ dkifer@cse.psu.edu \ giles@ist.psu.edu}
}

\maketitle

\begin{abstract}
   We present an end-to-end, multimodal, fully convolutional network for extracting semantic structures from document images. We consider document semantic structure extraction as a pixel-wise segmentation task, and propose a unified model that classifies pixels based not only on their visual appearance, as in the traditional page segmentation task, but also on the content of underlying text. Moreover, we propose an efficient synthetic document generation process that  we use to generate pretraining data for our network. Once the network is trained on a large set of synthetic documents, we fine-tune the network on unlabeled real documents using a semi-supervised approach. We systematically study the optimum network architecture and show that both our multimodal approach and the synthetic data pretraining significantly boost the performance.
\end{abstract}

\section{Introduction}
Document semantic structure extraction (DSSE) is an actively-researched area dedicated to understanding images of documents. The goal is to split a document image into regions of interest and to recognize the role of each region. It is usually done in two steps: the first step, often referred to as \textit{page segmentation}, is appearance-based and attempts to distinguish text regions from regions like figures, tables and line segments. The second step, often referred to as \textit{logical structure analysis}, is semantics-based and categorizes each region into semantically-relevant classes like paragraph and caption.

In this work, we propose a unified multimodal fully convolutional network (MFCN) that simultaneously identifies both \textit{appearance-based} and \textit{semantics-based} classes. It is a generalized page segmentation model that additionally performs fine-grained recognition on text regions: text regions are assigned specific labels based on their semantic functionality in the document. Our approach simplifies DSSE and better supports document image understanding.

\begin{figure}
\centering
\includegraphics[width=1\linewidth]{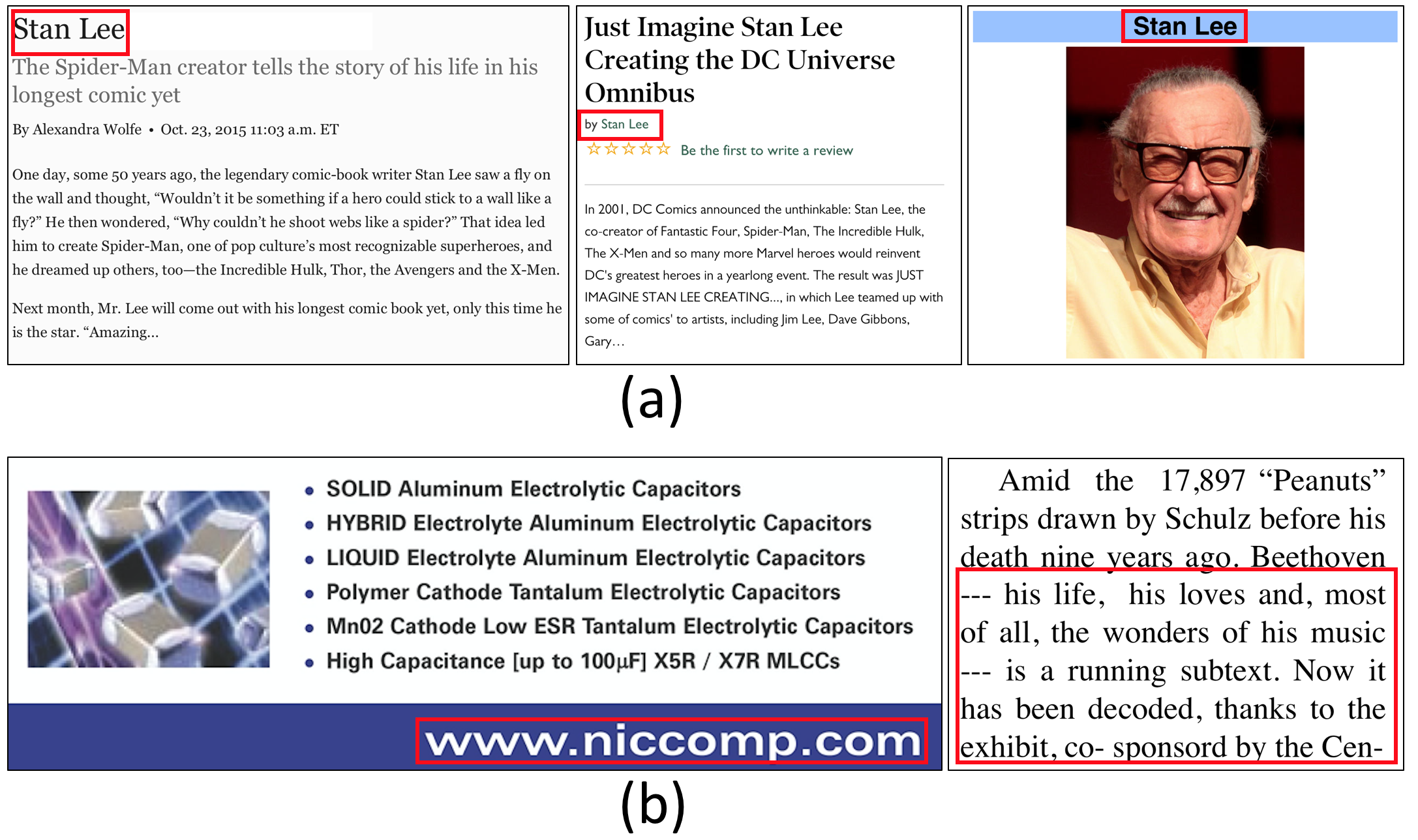}
\vspace{-0.15in}
\caption{(a) Examples that are difficult to identify if only based on text. The same name can be a title, an author or a figure caption. (b) Examples that are difficult to identify if only based on visual appearance. Text in the large font might be mislabeled as a section heading. Text with dashes might be mislabeled as a list.}
\label{figure:example:mislabel}
\vspace{-0.18in}
\end{figure}

We consider DSSE as a pixel-wise segmentation problem: each pixel is labeled as background, figure, table, paragraph, section heading, list, caption, etc. We show that our MFCN model trained in an end-to-end, pixels-to-pixels manner on document images exceeds the state-of-the-art significantly. It eliminates the need to design complex heuristic rules and extract hand-crafted features~\cite{lebourgeois1992fast,ha1995recursive,ha1995document,simon1997fast,amin2001page}.

In many cases, regions like section headings or captions can be visually identified. In Fig.~\ref{figure:example:mislabel}~(a), one can easily recognize the different roles of the same name. However, a robust DSSE system needs the semantic information of the text to disambiguate possible false identifications. For example, in Fig.~\ref{figure:example:mislabel}~(b), the text in the large font might look like section heading, but it does not function that way; the lines beginning with dashes might be mislabeled as a list.

\begin{figure*}
\centering
\includegraphics[width=1\linewidth]{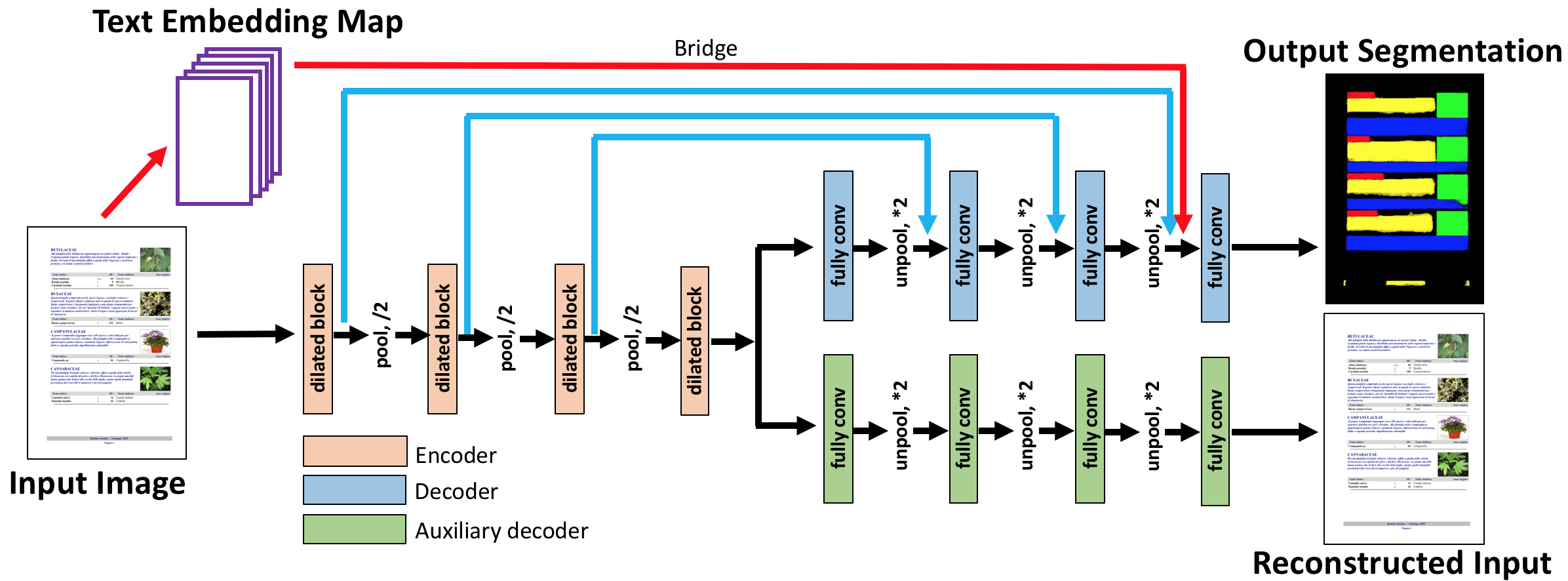}
\vspace{-0.25in}
\caption{The architecture of the proposed multimodal fully convolutional neural network. It consists of four parts: an encoder that learns a hierarchy of feature representations, a decoder that outputs segmentation masks, an auxiliary decoder for unsupervised reconstruction, and a bridge that merges visual representations and textual representations. The auxiliary decoder only exists during training.}
\label{figure:architecture}
\vspace{-0.18in}
\end{figure*}

To this end, our multimodal fully convolutional network is designed to leverage the textual information in the document as well. To incorporate textual information in a CNN-based architecture, we build a text embedding map and feed it to our MFCN. More specifically, we embed each sentence and map the embedding to the corresponding pixels where the sentence is represented in the document. Fig.~\ref{figure:architecture} summarizes the architecture of the proposed MFCN model. Our model consists of four parts: an encoder that learns a hierarchy of feature representations, a decoder that outputs segmentation masks, an auxiliary decoder for reconstruction during training, and a bridge that merges visual representations and textual representations. We assume that the document text has been pre-extracted. For document images this can be done with modern OCR engines~\cite{smith2007overview,ABBYY,Omnipage}.

One of the bottlenecks in training fully convolutional networks is the need for pixel-wise ground truth data. Previous document understanding datasets~\cite{liang1997uw,sauvola1999mediateam,todoran2005uva,antonacopoulos2009realistic} are limited by both their small size and the lack of fine-grained semantic labels such as section headings, lists, or figure and table captions. To address these issues, we propose an efficient synthetic document generation process and use it to generate large-scale pretraining data for our network. Furthermore, we propose two unsupervised tasks for better generalization to real documents: reconstruction and consistency tasks. The former enables better representation learning by reconstructing the input image, whereas the latter encourages pixels belonging to the same regions have similar representation.

Our main contributions are summarized as follows:
\begin{itemize}
	\item We propose an end-to-end, unified network to address document semantic structure extraction. Unlike previous two-step processes, we simultaneously identify both \textit{appearance-based} and \textit{semantics-based} classes. 
	\item Our network supports both supervised training on image and text of documents, as well as unsupervised auxiliary training for better representation learning.
	\item We propose a synthetic data generation process and use it to synthesize a large-scale dataset for training the supervised part of our deep MFCN model.
\end{itemize}

\section{Background}\label{section:related}
\textbf{Page Segmentation.} Most earlier works on page segmentation~\cite{lebourgeois1992fast,ha1995recursive,ha1995document,simon1997fast,amin2001page,shilman2005learning}
 fall into two categories: bottom-up and top-down approaches. Bottom-up approaches~\cite{lebourgeois1992fast,simon1997fast,amin2001page} first detect words based on local features (white/black pixels or connected components), then sequentially group words into text lines and paragraphs. However, such approaches suffer from the identification and grouping of connected components being time-consuming. Top-down approaches~\cite{ha1995recursive,ha1995document} iteratively split a page into columns, blocks, text lines and words. With both of these approaches it is difficult to correctly segment documents with complex layout, for example a document with non-rectangular figures~\cite{mao2003document}.
 
With recent advances in deep convolutional neural networks, several neural-based models have been proposed. Chen et al.~\cite{chen2015page} applied a convolutional auto-encoder to learn features from cropped document image patches, then use these features to train a SVM~\cite{cortes1995support} classifier. Vo et al.~\cite{vo2016dense} proposed using FCN to detect lines in handwritten document images. However, these methods are strictly restricted to visual cues, and thus are not able to discover the semantic meaning of the underlying text.

\textbf{Logical Structure Analysis.} Logical structure is defined as a hierarchy of logical components in documents, such as section headings, paragraphs and lists~\cite{mao2003document}. Early work in logical structure discovery~\cite{fisher1991logical,krishnamoorthy1993syntactic,ingold1991top,conway1993page} focused on using a set of heuristic rules based on the location, font and text of each sentence. Shilman et al.~\cite{shilman2005learning} modeled document layout as a grammar and used machine learning to minimize the cost of a invalid parsing. Luong et al.~\cite{luong2012logical} proposed using a conditional
random fields model to jointly label each sentence based on several hand-crafted features. However, the performance of these methods is limited by their reliance on hand-crafted features, which cannot capture the highly semantic context.
 
\textbf{Semantic Segmentation.} Large-scale annotations~\cite{lin2014microsoft} and the development of deep neural network approaches such as the fully convolutional network (FCN)~\cite{long2015fully} have led to rapid improvement of the accuracy of semantic segmentation \cite{chen2014semantic,papandreou2015weakly,noh2015learning,yu2015multi}. However, the originally proposed FCN model has several limitations, such as ignoring small objects and mislabeling large objects due to the fixed receptive field size. To address this issue, Noh et al.~\cite{noh2015learning} proposed using unpooling, a technique that reuses the pooled ``location'' at the up-sampling stage.  Pinheiro et al.~\cite{pinheiro2016learning} attempted to use skip connections to refine segmentation boundaries. Our model addresses this issue by using a dilated block, inspired by dilated convolutions~\cite{yu2015multi} and recent work \cite{szegedy2015going,he2015deep} that groups several layers together . We further investigate the effectiveness of different approaches to optimize our network architecture.

Collecting pixel-wise annotations for thousands or millions of images requires massive labor and cost. To this end, several methods~\cite{papandreou2015weakly,zhang2015weakly,lu2016learning} have been proposed to harness weak annotations (bounding-box level or image level annotations) in neural network training. Our consistency loss relies on similar intuition but does not require a ``class label'' for each bounding box.

\textbf{Unsupervised Learning.} Several methods have been proposed to use unsupervised learning to improve supervised learning tasks. Mairal et al.~\cite{mairal2009supervised} proposed a sparse coding method that learns sparse local features by sparsity-constrained reconstruction loss functions. Zhao et al.~\cite{zhao2015stacked} proposed a Stacked What-Where Auto-Encoder that uses unpooling during reconstruction. By injecting noise into the input and the middle features, a denoising auto-encoder~\cite{vincent2008extracting} can learn robust filters that recover uncorrupted input. The main focus in unsupervised learning has been image-level classification and generative approaches, whereas in this paper we explore the potential of such methods for pixel-wise semantic segmentation.

Wen et al.~\cite{wen2016discriminative} recently proposed a center loss  that encourages data samples with the same label to have a similar visual representation. Similarly, we introduce an intra-class consistency constraint. However, the ``center'' for each class in their loss is determined by data samples across the whole dataset, while in our case the ``center'' is locally determined by pixels within the same region in each image.

\textbf{Language and Vision.} Several joint learning tasks such as image captioning~\cite{donahue2015long,karpathy2015deep}, visual question answering~\cite{antol2015vqa,gao2015you,malinowski2015ask}, and one-shot learning~\cite{frome2013devise,socher2013zero,changpinyo2016synthesized} have demonstrated the significant impact of using textual and visual representations in a joint framework. Our work is unique in that we use textual embedding \textit{directly} for a segmentation task for the first time, and we show that our approach improves the results of traditional segmentation approaches that only use visual cues.

\section{Method}\label{section:methods}
Our method does supervised training for pixel-wise segmentation with a specialized multimodal fully convolutional network that uses a text embedding map jointly with the visual cues. Moreover, our MFCN architecture also supports two unsupervised learning tasks to improve the learned document representation: a reconstruction task based on an auxiliary decoder and a consistency task evaluated in the main decoder branch along with the per-pixel segmentation loss.

\subsection{Multimodal Fully Convolutional Network}\label{subsection:model}
As shown in Fig.~\ref{figure:architecture}, our MFCN model has four parts: an encoder, two decoders and a bridge. The encoder and decoder parts roughly follow the architecture guidelines set forth by Noh et al.~\cite{noh2015learning}. However, several changes have been made to better address document segmentation.

First, we observe that several semantic-based classes such as section heading and caption usually occupy relatively small areas. Moreover, correctly identifying certain regions often relies on small visual cues, like lists being identified by small bullets or numbers in front of each item. This suggests that low-level features need to be used. However, because max-pooling naturally loses information during downsampling, FCN often performs poorly for small objects. Long et al.~\cite{long2015fully} attempt to avoid this problem using skip connections. However, simply averaging independent predictions based on features at different scales does not provide a satisfying solution. Low-level representations, limited by the local receptive field, are not aware of object-level semantic information; on the other hand, high-level features are not necessarily aligned consistently with object boundaries because CNN models are invariant to translation. We propose an alternative skip connection implementation, illustrated by the blue arrows in Fig.~\ref{figure:architecture}, similar to that used in the independent work \textit{SharpMask}~\cite{pinheiro2016learning}. However, they use bilinear upsampling after skip connection while we use unpooling to preserve more spatial information.

\begin{figure}
\centering
\includegraphics[width=1\linewidth]{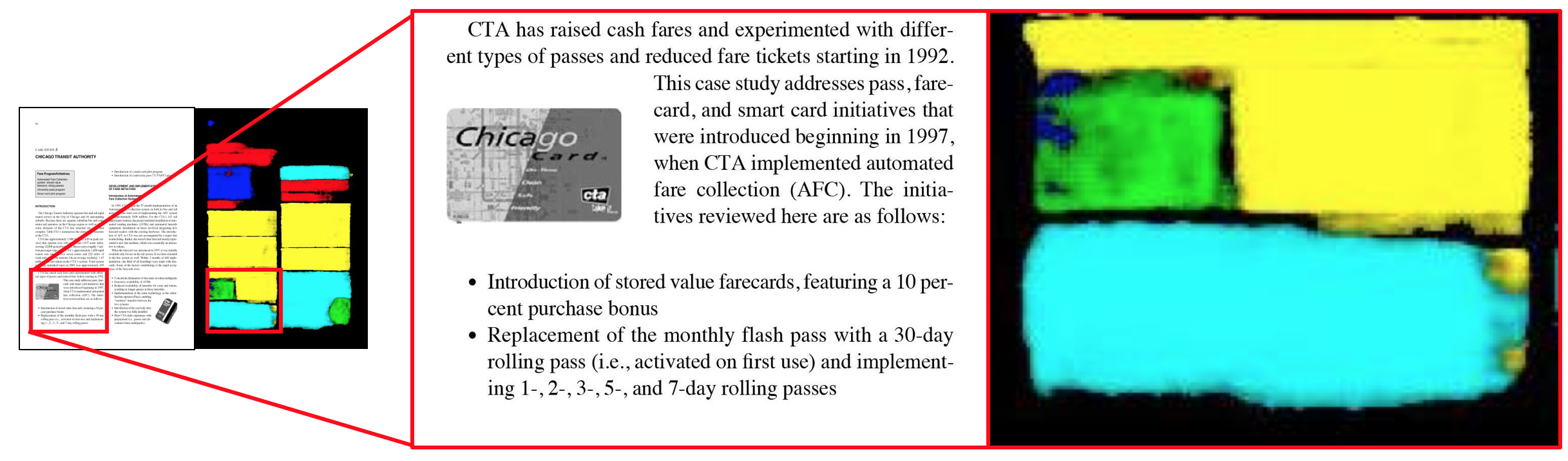}
\vspace{-0.22in}
\caption{A cropped document image and its segmentation mask generated by our model. Note that the top-right corner of the list is yellow instead of cyan, indicating that it has been mislabeled as a paragraph.}
\label{figure:example:dilated}
\vspace{-0.15in}
\end{figure}

We also notice that broader context information is needed to identify certain objects. For an instance, it is often difficult to tell the difference between a list and several paragraphs by only looking at parts of them. In Fig.~\ref{figure:example:dilated}, to correctly segment the right part of the list, the receptive fields must be large enough to capture the bullets on the left. Inspired by the Inception architecture~\cite{szegedy2015going} and dilated convolution~\cite{yu2015multi}, we propose a dilated convolution block, which is illustrated in Fig.~\ref{figure:dilated}~(left). Each dilated convolution block consists of 5 dilated convolutions with a $3\times3$ kernel size and a dilation $d = 1, 2, 4, 8, 16$.

\subsection{Text Embedding Map}\label{subsection:embedding}
Traditional image semantic segmentation models learn the semantic meanings of objects from a visual perspective. Our task, however, also requires understanding the text in images from a linguistic perspective. Therefore, we build a text embedding map and feed it to our multimodal model to make use of both visual and textual representations.

We treat a sentence as the minimum unit that conveys certain semantic meanings, and represent it using a low-dimensional vector. Our sentence embedding is built by averaging embeddings for individual words. This is a simple yet effective method that has been shown to be useful in many applications, including sentiment analysis~\cite{iyyer2015deep} and text classification~\cite{joulin2016bag}. Using such embeddings, we create a text embedding map as follows: for each pixel inside the area of a sentence, we use the corresponding sentence embedding as the input. Pixels that belong to the same sentence thus share the same embedding. Pixels that do not belong to any sentences will be filled with zero vectors. For a document image of size $H \times W$, this process results in an embedding map of size $N \times H \times W$ if the learned sentence embeddings are $N$-dimensional vectors. The embedding map is later concatenated with a feature response along the number-of-channel dimensions (see Fig.~\ref{figure:architecture}).

\begin{figure}
\centering
\includegraphics[width=0.45\linewidth]{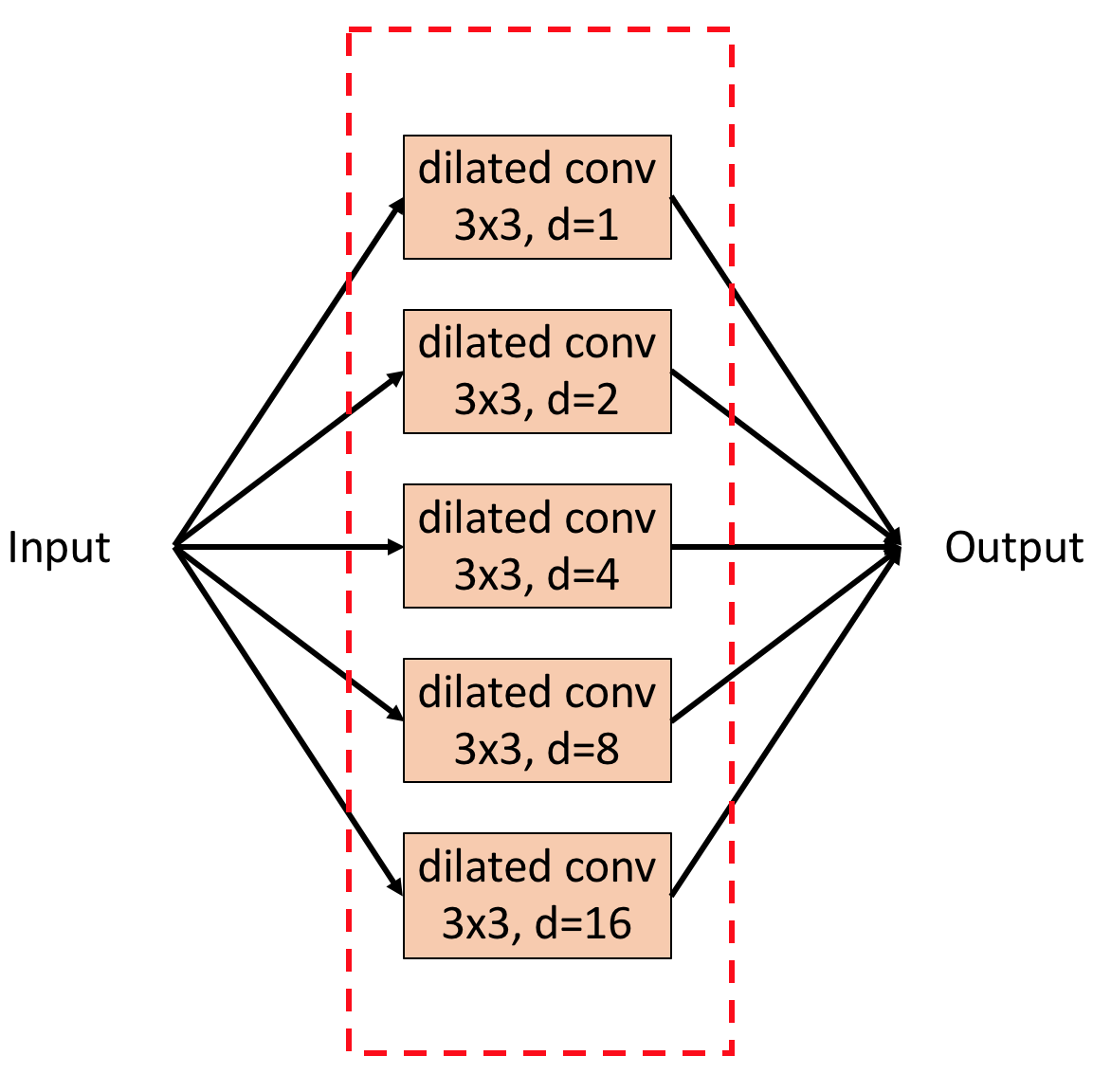}
\hspace{1em}
\includegraphics[width=0.35\linewidth]{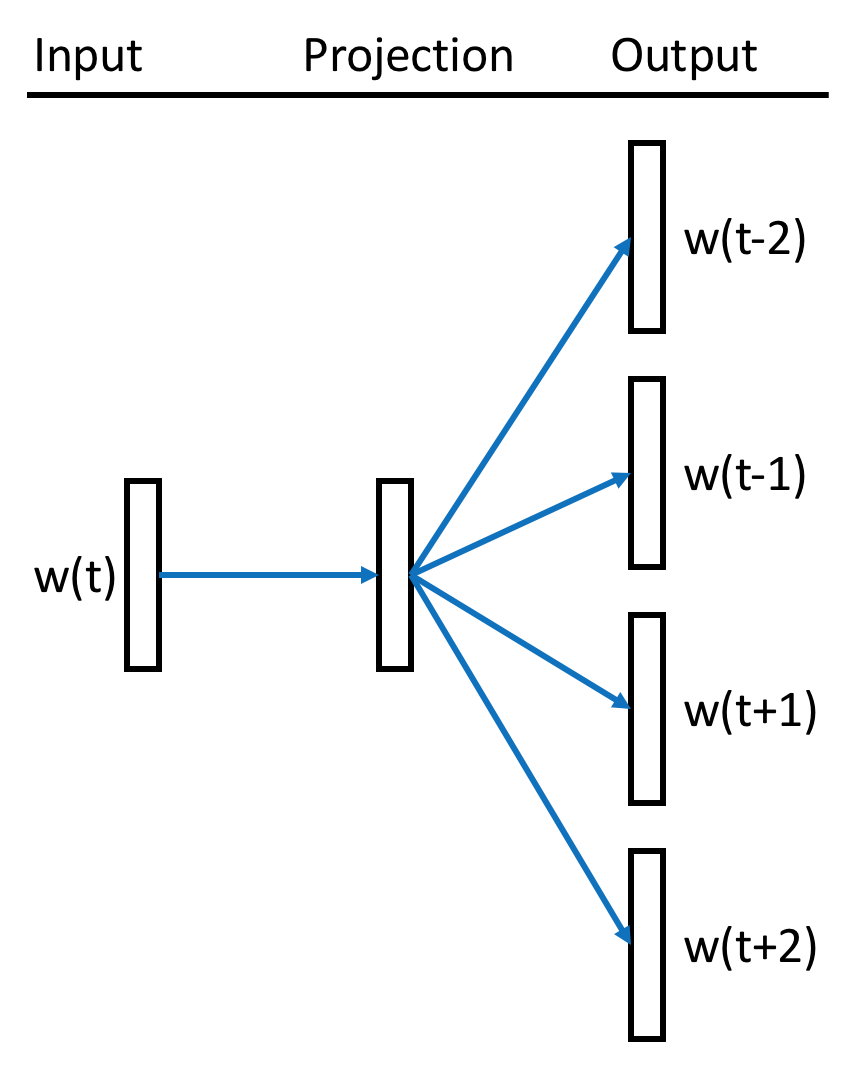}
\vspace{-0.12in}
\caption{Left: A dilated block that contains 5 dilated convolutional layers with different dilation $d$. Batch-Normalization and non-linearity are not shown for brevity. Right: The skip-gram model for word embeddings.}
\label{figure:dilated}
\vspace{-0.15in}
\end{figure}

Specifically, our word embedding is learned using the skip-gram model~\cite{mikolov2013efficient,mikolov2013distributed}. Fig.~\ref{figure:dilated}~(right) shows the basic diagram. Let $V$ be the number of words in a vocabulary and $w$ be a $V$-dimensional one-hot vector representing a word. The training objective is to find a $N$-dimensional ($N \ll V$) vector representation for each word that is useful for predicting the neighboring words. More formally, given a sequence of words $[w_1, w_2, \cdots, w_T]$, we maximize the average log probability
\begin{align}
	\frac{1}{T} \sum_{t=1}^T \sum_{-C \leq j \leq C, j \neq 0} \text{log} P(w_{t+j} | w_t)
\end{align}
\noindent where $T$ is the length of the sequence and $C$ is the size of the context window. The probability of outputting a word $w_o$ given an input word $w_i$ is defined using softmax:
\begin{align}
	P(w_o | w_i) = \frac{\text{exp} ({v_{w_o}^{'}}^{\top} v_{w_i})}{\sum_{w=1}^V \text{exp} ({v_{w}^{'}}^{\top} v_{w_i})}
\end{align}
\noindent where $v_w$ and $v_w^{'}$ are the ``input'' and ``output'' $N$-dimensional vector representations of $w$.

\subsection{Unsupervised Tasks}\label{subsection:unsupervised}
Although our synthetic documents (Sec.~\ref{section:data}) provide a large amount of labeled data for training, they are limited in the variations of their layouts. To this end, we define two unsupervised loss functions to make use of real documents and to encourage better representation learning. 

\textbf{Reconstruction Task.} It has been shown that reconstruction can help learning better representations and therefore improves performance for supervised tasks~\cite{zhao2015stacked,zhangaugmenting}. We thus introduce a second decoder pathway (Fig.~\ref{figure:architecture} - axillary decoder), denoted as $D_{rec}$, and define a reconstruction loss at intermediate features. This auxiliary decoder only exists during the training phase.

Let $a_l, l = 1, 2, \cdots L$ be the activations of the $l^{th}$ layer of the encoder, and $a_0$ be the input image. For a feed-forward convolutional network, $a_l$ is a feature map of size $C_l \times H_l \times W_l$. Our auxiliary decoder $D_{rec}$ attempts to reconstruct a hierarchy of feature maps $\{\tilde{a}_l\}$. Reconstruction loss $L_{rec}^{(l)}$ for a specific $l$ is therefore defined as
\begin{align}
	L_{rec}^{(l)} = \frac{1}{C_l H_l W_l} \norm{a_l - \tilde{a}_l}_2^2, \quad l = 0, 1, 2, \cdots L
\end{align}

\textbf{Consistency Task.} Pixel-wise annotations are labor-intensive to obtain, however it is relatively easy to get a set of bounding boxes for detected objects in a document. For documents in PDF format, one can find bounding boxes by analyzing the rendering commands in the PDF files (See our supplementary document for typical examples). Even if their labels remain unknown, these bounding boxes are still beneficial: they provide knowledge of which parts of a document belongs to the same objects and thus should not be segmented into different fragments.

By building on the intuition that regions belonging to same objects should have similar feature representations, we define the consistency task loss $L_{cons}$ as follows. Let $p_{(i,j)}$ $(i=1, 2, \cdots H, j=1, 2, \cdots W)$ be activations at location $(i, j)$ in a feature map of size $C \times H \times W$, and $b$ be the rectangular area in a bounding box. Let each rectangular area $b$ is of size $H_b \times W_b$. Then, for each $b \in B$, $L_{cons}$ will be given by
\begin{align}
	L_{cons} &= \frac{1}{H_b W_b} \sum_{(i,j) \in b} \norm{p_{(i,j)} - p^{(b)}}_2^2\\
	p^{(b)} &= \frac{1}{H_b W_b} \sum_{(i,j) \in b} p_{(i,j)}
\end{align} 
Minimizing consistency loss $L_{cons}$ encourages intra-region consistency.

The consistency loss $L_{cons}$ is differentiable and can be optimized using stochastic gradient descent. The gradient of $L_{cons}$ with respect to $p_{(i,j)}$ is
\begin{align}
	\frac{\partial L_{cons}}{\partial p_{(i,j)}} = 
		&\frac{2}{H_b^2 W_b^2}
		(p_{(i,j)} - p^{(b)}) (H_b W_b - 1) + \nonumber \\
		&\frac{2}{H_b^2 W_b^2}
		\sum_{\mathclap{\substack{(u,v) \in b\\
			(u,v) \neq (i,j)}}}
		(p^{(b)} - p_{(u,v)})
\end{align}

\noindent since $H_b W_b \gg 1$, for efficiency it can be approximated by:
\begin{align}
	\frac{\partial L_{cons}}{\partial p_{(i,j)}} \approx 
		&\frac{2}{H_b W_b} \left(p_{(i,j)} - p^{(b)}\right).
\end{align}
We use the unsupervised consistency loss, $L_{cons}$, as a loss layer, that is evaluated at the main decoder branch (blue branch in Fig.~\ref{figure:architecture}) along with supervised segmentation loss.

\section{Synthetic Document Data}\label{section:data}
Since our MFCN aims to generate a segmentation mask of the whole document image, pixel-wise annotations are required for the supervised task. While there are several publicly available datasets for page segmentation~\cite{sauvola1999mediateam,todoran2005uva,antonacopoulos2009realistic}, there are only a few hundred to a few thousand pages in each. Furthermore, the types of labels are limited, for example to text, figure and table, however our goal is to perform a much more granular segmentation.

To address these issues, we created a synthetic data engine, capable of generating large-scale, pixel-wise annotated documents.

\begin{figure*}
\centering
{
\includegraphics[width=1\linewidth]{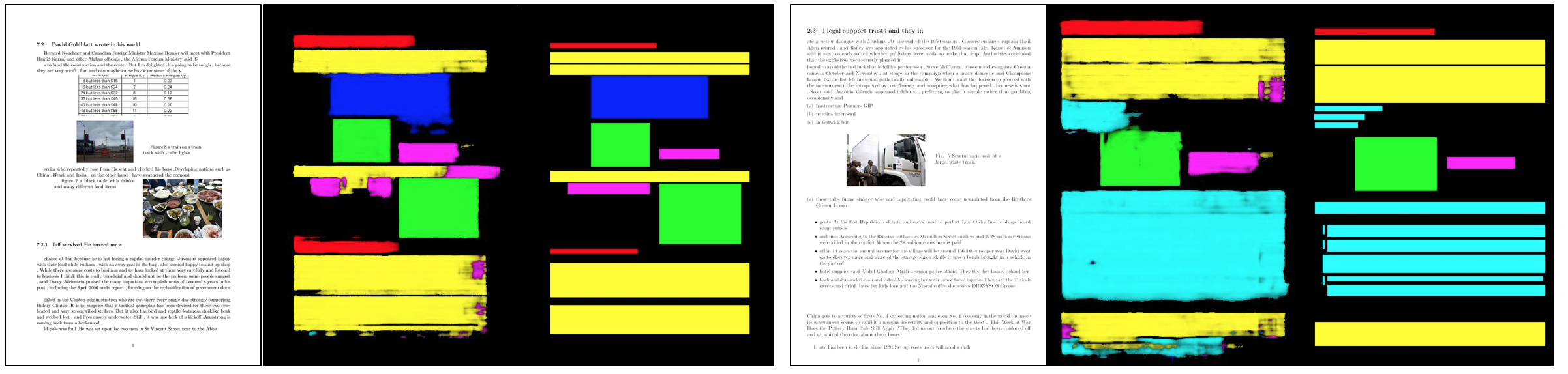}
}
\vspace{-0.25in}
\caption{Example synthetic documents, raw segmentations and results after optional post-processing (Sec.~\ref{section:implement}). Segmentation label colors are: \colorbox{green}{\strut \bf figure}, \colorbox{blue}{\strut \textcolor{white}{\bf table}}, \colorbox{red}{\strut \textcolor{white}{\bf section heading}}, \colorbox{mymagenta}{\strut \textcolor{white}{\bf caption}}, \colorbox{mycyan}{\strut \bf list} and \colorbox{yellow}{\strut \bf paragraph}.}
\label{figure:synthetic}
\vspace{-0.18in}
\end{figure*}

Our synthetic document engine uses two methods to generate documents. The first produces completely automated and random layout of partial data scraped from the web. More specifically, we generate LaTeX source files in which paragraphs, figures, tables, captions, section headings and lists are randomly arranged to make up single, double, or triple-column PDFs. Candidate figures include academic-style figures and graphic drawings downloaded using web image search, and natural images from MS COCO~\cite{lin2014microsoft}, which associates each image with several captions. Candidate tables are downloaded using web image search. Various queries are used to increase the diversity of downloaded tables. Since our MFCN model relies on the semantic meaning of text to make prediction, the content of text regions (paragraph, section heading, list, caption) must be carefully selected:
\begin{itemize}
	\item For paragraphs, we randomly sample sentences from a 2016 English Wikipedia dump~\cite{Wikipedia}.
	\item For section headings, we only sample sentences and phrases that are section or subsection headings in the ``Contents'' block in a Wikipedia page.
	\item For lists, we ensure that all items in a list come from the same Wikipedia page.
	\item For captions, we either use the associated caption (for images from MS COCO) or the title of the image in web image search, which can be found in the span with class name ``irc\_pt''.
\end{itemize}

To further increase the complexity of the generated document layouts, we collected and labeled 271 documents with varied, complicated layouts. We then randomly replaced each element with a standalone paragraph, figure, table, caption, section heading or list generated as stated above.

In total, our synthetic dataset contains 135,000 document images. Examples of our synthetic documents are shown in Fig.~\ref{figure:synthetic}. Please refer to our supplementary document for more examples of synthetic documents and individual elements used in the generation process.

\section{Implementation Details}\label{section:implement}
Fig.~\ref{figure:architecture} summarizes the architecture of our model. The auxiliary decoder only exists in the training phase. All convolutional layers have a $3 \times 3$ kernel size and a stride of $1$. The pooling (in the encoders) and unpooling (in the decoders) have a kernel size of $2 \times 2$. We adopt batch normalization~\cite{ioffe2015batch} immediately after each convolution and before all non-linear functions.

We perform per-channel mean subtraction and resize each input image so that its longer side is less than 384 pixels. No other pre-processing is applied. We use Adadelta~\cite{zeiler2012adadelta} with a mini-batch size of 2. During semi-supervised training, mini-batches of synthetic and real documents are used alternatively. For synthetic documents, both per-pixel classification loss and the unsupervised losses are active at back-propagation, while for real documents, only the unsupervised losses are active. Since the labels are unbalanced (e.g. the area of paragraphs is much larger than that of caption), class weights for the per-pixel classification loss are set differently according to the total number of pixels in each class in the training set. 

For text embedding, we represent each word as a $128$-dimensional vector and train a skip-gram model on the 2016 English Wikipedia dump~\cite{Wikipedia}. Embeddings for out-of-dictionary words are obtained following Bojanowski et al.~\cite{bojanowski2016enriching}. We use Tesseract~\cite{smith2007overview} as our OCR engine.

\textbf{Post-processing.} We apply an optional post-processing step as a cleanup strategy for segment masks. For documents \textit{in PDF format}, we obtain a set of candidate bounding boxes by analyzing the PDF format to find element boxes. We then refine the segmentation masks by first calculating the average class probability for pixels belonging to the same box, followed by assigning the most likely label to these pixels. 

\section{Experiments}\label{section:experiments}
We used three datasets for evaluations: ICDAR2015~\cite{antonacopoulos2009realistic}, SectLabel~\cite{luong2012logical} and our new dataset named DSSE-200. ICDAR2015~\cite{antonacopoulos2009realistic} is a dataset used in the biennial ICDAR page segmentation competitions~\cite{antonacopoulos2015icdar2015} focusing more on appearance-based regions. The evaluation set of ICDAR2015 consists of 70 sampled pages from contemporary magazines and technical articles. SectLabel~\cite{luong2012logical} consists of 40 academic papers with 347 pages in the field of computer science. Each text line in these papers is manually assigned a semantics-based label such as text, section heading or list item. In addition to these two datasets, we introduce DSSE-200\footnote{\url{http://personal.psu.edu/xuy111/projects/cvpr2017_doc.html}.}, which provides both appearance-based and semantics-based labels. DSSE-200 contains 200 pages from magazines and academic papers. Regions in a page are assigned labels from the following dictionary: figure, table, section, caption, list and paragraph. Note that DSSE-200 has a more granular segmentation than previously released benchmark datasets.

\begin{figure*}
\centering
{
\includegraphics[width=1\linewidth]{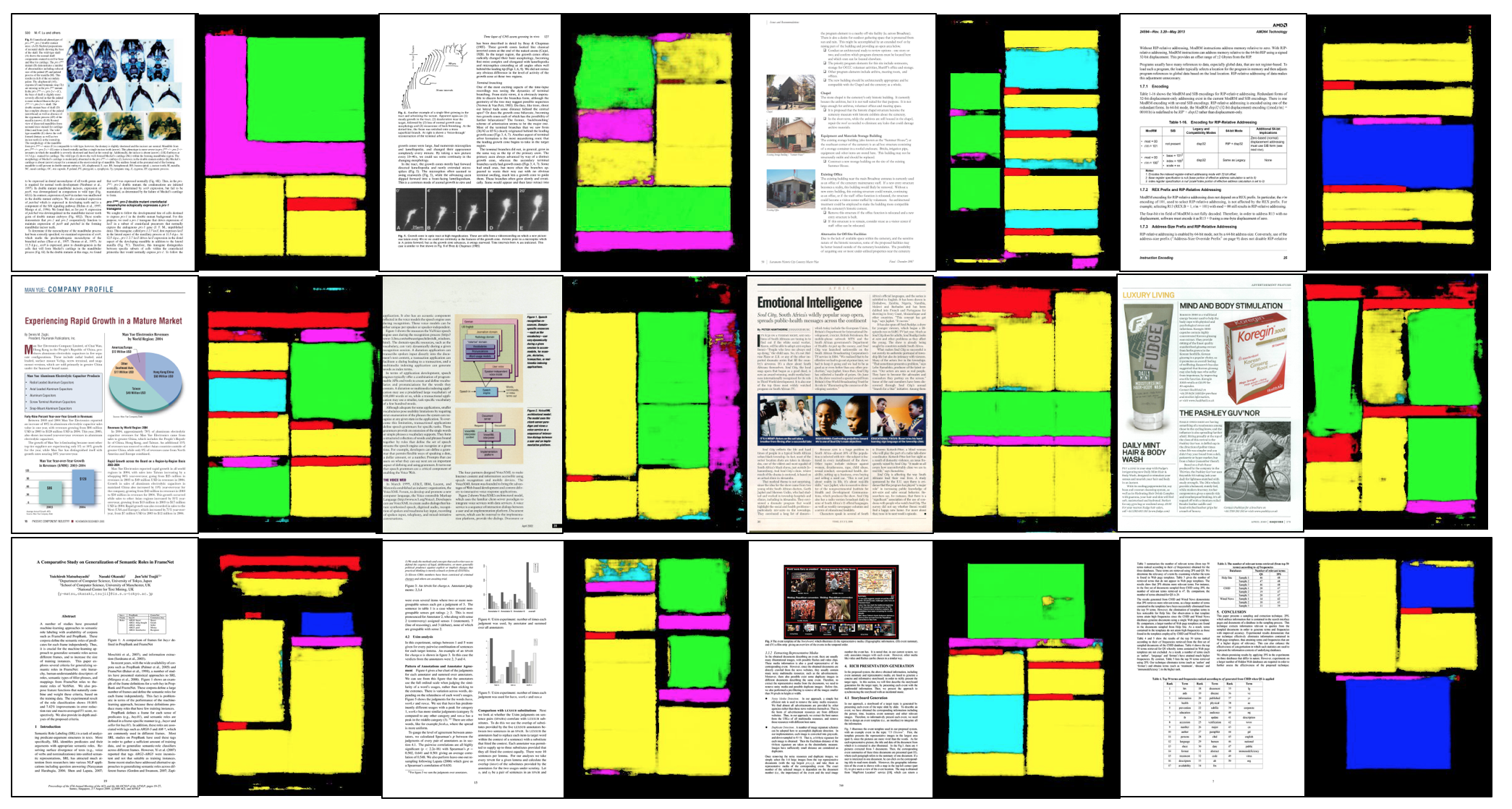}
}
\vspace{-0.20in}
\caption{Example real documents and their corresponding segmentation. Top: DSSE-200. Middle: ICDAR2015. Bottom: SectLabel. Since these documents are not in PDF format, the simple post-processing in Sec.~\ref{section:implement} can not be applied. One may consider exploiting a CRF~\cite{chen2014semantic} to refine the segmentation, but that is beyond the main focus of this paper. Segmentation label colors are: \colorbox{green}{\strut \bf figure}, \colorbox{blue}{\strut \textcolor{white}{\bf table}}, \colorbox{red}{\strut \textcolor{white}{\bf section heading}}, \colorbox{mymagenta}{\strut \textcolor{white}{\bf caption}}, \colorbox{mycyan}{\strut \bf list} and \colorbox{yellow}{\strut \bf paragraph}. }
\label{figure:datasets}
\vspace{-0.00in}
\end{figure*}

The performance is measured in terms of pixel-wise intersection-over-union (IoU), which is standard in semantic segmentation tasks. We optimize the architecture of our MFCN model based on the DSSE-200 dataset since it contains both appearance-based and semantics-based labels. Sec.~\ref{subsection:exp:comparison} compares our results to state-of-the-art methods on the ICDAR2015 and SectLabel datasets.

\subsection{Ablation Experiment on Model Architecture}
We first systematically evaluate the effectiveness of different network architectures. Results are shown in Table~\ref{table:ablation}. Note that these results do not incorporate textual information or unsupervised learning tasks. The purpose of this experiment is to find the best ``base'' architecture to be used in the following experiments. All models are trained from scratch and evaluated on the DSSE-200 dataset.

\begin{table*}
\begin{center}
\begin{tabular}{| c || c c c || c c c c c c c || c |} \hline
 Model\# & dilation & upsampling & skip & bkg & figure & table & section & caption & list & paragraph & mean \\ \hline \hline
 1 & 1 & bilinear & no & 80.3 & 75.4 & 62.7 & 50.0 & 33.8 & 57.3 & 70.4 & 61.4 \\
 2 & 1 & bilinear & yes & 82.1 & 76.7 & 74.4 & 51.8 & 42.4 & 58.7 & 74.4 & 65.4 \\
 3 & 1 & unpooling & yes & 84.1 & 81.2 & 77.6 & 54.6 & 60.3 & 65.9 & 74.8 & 71.2 \\
 4 & 8 & unpooling & yes & 83.9 & 74.9 & 69.7 & 57.2 & 60.2 & 64.6 & 76.1 & 69.5 \\
 5 & block & unpooling & yes & \textbf{84.6} & \textbf{83.3} & \textbf{79.4} & \textbf{58.3} & \textbf{61.0} & \textbf{66.7} & \textbf{77.1} & \textbf{73.0} \\ \hline
\end{tabular}
\end{center}
\vspace{-0.20in}
\caption{Ablation experiments on DSSE-200 dataset. The architecture of each model is characterized by the dilation in convolution layers, the way of upsampling and the use of skip connection. IoU scores (\%) are reported.}
\label{table:ablation}
\vspace{-0.15in}
\end{table*}

As a simple baseline (Table~\ref{table:ablation} Model1), we train a plain encoder-decoder style model for document segmentation. It consists of a feed-forward convolutional network as an encoder, and a decoder implemented by a fully convolutional network. Upsampling is done by bilinear interpolation. This model achieves a mean IoU of 61.4\%.

Next, we add skip connections to the model, resulting in Model2. Note that this model is similar to the \textit{SharpMask} model. We observe a mean IoU of 65.4\%, 4\% better than the base model. The improvements are even more significant for small objects like captions.

We further evaluate the effectiveness of replacing bilinear upsampling with unpooling, giving Model3. All upsampling layers in Model2 are replaced by unpooling while other parts are kept unchanged. Doing so results in a significant improvement for mean IoU (65.4\% vs. 71.2\%). This suggests that the pooled index should not be discarded during decoding. These indexes are helpful to disambiguate the location information when constructing the segmentation mask in the decoder.

Finally, we investigate the use of dilated convolutions. Model3 is equivalent to using dilated convolution when $d=1$. Model4 sets $d=8$ while Model5 uses the dilated block illustrated in Fig.~\ref{figure:dilated} (left). The number of output channels are adjusted such that the total number of parameters are similar. Comparing the results for these three models, we can see that the IoU of Model4 for each class is on par with or worse than Model3, while Model5 is better than both Model3 and Model4 for all classes.

\begin{table*}
\begin{center}
\begin{tabular}{| l || c c || ccccccc || c |} \hline
	base & dataset & text & bkg & figure & table & section & caption & list & para. & mean \\ \hline \hline
	Model5 & D & none & \textbf{84.6} & 83.3 & \textbf{79.4} & 58.3 & 61.0 & 66.7 & 77.1 & 73.0 \\
	Model5 & D & extract & 83.9 & \textbf{83.7} & 79.7 & \textbf{59.4} & \textbf{61.1} & \textbf{68.4} & \textbf{79.3} & \textbf{73.3}\\ \hline \hline
	Model5 & S & none & 87.7 & 83.1 & 84.3 & 70.8 & 70.9 & 82.3 & 83.1 & 79.6 \\
	Model5 & S & extract & 88.8 & 85.4 & 86.6 & 73.1 & 71.2 & 83.6 & 87.2 & 82.2\\
	Model5 & S & real & \textbf{91.2} & \textbf{90.3} & \textbf{89.0} & \textbf{78.4} & \textbf{75.3} & \textbf{87.5} & \textbf{89.6} & \textbf{86.0}\\ \hline
\end{tabular}
\end{center}
\vspace{-0.25in}
\caption{IoU scores (\%) on the DSSE-200 (D) and synthetic dataset (S) using text embedding map. On synthetic dataset, we further investigate the effects of using extracted text versus real text when building the text embedding map.}
\label{table:text}
\vspace{-0.15in}
\end{table*}

\subsection{Adding Textual Information}\label{subsection:exp:text}
We now investigate the importance of textual information in our multimodal model. We take the best architecture, Model5, as our vision-only model, and incorporate a text embedding map via a bridge module depicted in Fig.~\ref{figure:architecture}. This combined model is fine-tuned on our synthetic documents. As shown in Table~\ref{table:text}, using text as well improves the performance for \textit{textual} classes. The accuracy for section heading, caption, list and paragraph is boosted by 1.1\%, 0.1\%, 1.7\% and 2.2\%, respectively. 

We rely on existing OCR engines~\cite{smith2007overview} to extract text, but they are not always reliable for scanned documents of low quality. To quantitatively analyze the effects of using extracted text, we compare the performance of using extracted text versus real text. The comparison is conducted on a subset of our synthetic dataset (200 images), since ground-truth text is naturally available. As shown in Table~\ref{table:text}, using real text leads to a remarkable improvement (6.4\%) for mean IoU, suggesting the effectiveness of incorporating textual information. Using OCR extracted text is not as effective, but still results in 2.6\% improvement. It is better than the 0.3\% improvement on DSSE-200 dataset; we attribute this to our synthetic data not being as complicated as DSSE-200, so extracting text becomes easier. 

\begin{table}
\begin{center}
\begin{tabular}{| c | c | c | c | c |} \hline
	& $L_{cls}$ & $L_{rec}$ & $L_{cons}$ & $L_{rec+con}$ \\ \hline
	mean & 73.3 & 73.9 & 75.4 & 75.9 \\ \hline
\end{tabular}
\end{center}
\vspace{-0.25in}
\caption{IoU scores (\%) when using different training objectives on DSSE-200 dataset. \textit{cls}: pixel-wise classification task, \textit{rec}: reconstruction task and \textit{cons}: consistency task.}
\label{table:unsupervised}
\begin{center}
\begin{tabular}{| l | c | c |} \hline
	Methods & non-text & text \\ \hline \hline
	Leptonica~\cite{bloomberg2007document} & 84.7 & 86.8 \\
	Bukhari et al.~\cite{bukhari2011improved} & 90.6 & 90.3 \\
	Ours (binary) & \textbf{94.5} & \textbf{91.0} \\ \hline \hline
	Methods & figure & text \\ \hline \hline
	Fernandez et al.~\cite{fernandez2012document} & 70.1 & 85.8 \\
	Ours (binary) & \textbf{77.1} & \textbf{91.0} \\ \hline
\end{tabular}
\end{center}
\vspace{-0.25in}
\caption{IoU scores (\%) for page segmentation on the ICDAR2015 dataset. For comparison purpose, only IoU scores for non-text, text and figure are shown. However our model can make fine-grained predictions as well.}
\label{table:comparison}
\begin{center}
\begin{tabular}{| l | c | c | c | c |} \hline
	Methods & section & caption & list & para. \\ \hline \hline
	Luong et al.~\cite{luong2012logical} & 0.916 & 0.781 & 0.712 & \textbf{0.969} \\
	Ours & \textbf{0.919} & \textbf{0.893} & \textbf{0.793} & \textbf{0.969}\\ \hline
\end{tabular}
\end{center}
\vspace{-0.25in}
\caption{F1 scores on the SectLabel dataset. Note that our model can also identify non-text classes such as figures and tables.}
\label{table:comparison:sectlabel}
\vspace{-0.18in}
\end{table}

\subsection{Unsupervised Learning Tasks}
Here, we examine how the proposed two unsupervised learning tasks --- reconstruction and consistency tasks --- can complement the pixel-wise classification during training. We take the best model in Sec.~\ref{subsection:exp:text}, and only change the training objectives. Our model is then fine-tuned in a semi-supervised manner as described in Sec.~\ref{section:implement}. The results are shown in Table~\ref{table:unsupervised}. Adding the reconstruction task slightly improves the mean IoU by 0.6\%, while adding the consistency task leads to a boost of 1.9\%. These results justify our hypothesis that harnessing region information is beneficial. Combining both tasks results in a mean IoU of 75.9\%.

\subsection{Comparisons with Prior Art}\label{subsection:exp:comparison}
Table~\ref{table:comparison} and \ref{table:comparison:sectlabel} present comparisons with several methods that have previously reported performance on the ICDAR2015 and SectLabel datasets. It is worth emphasizing that our MFCN model simultaneously predicts both appearance-based and semantics-based classes while other methods can not.

\textbf{Comparisons on ICDAR2015 dataset} (Table~\ref{table:comparison}). Previous pixel-wise page segmentation models usually solve a binary segmentation problem and do not make predictions for fine-grained classes. For fair comparison, we change the number of output channels of the last layer to 3 (background, figure and text) and fine-tune this last layer. Our binary MFCN model achieves 94.5\%, 91.0\% and 77.1\% IoU scores for non-text (background and figure), text and figure regions, outperforming other models. 

\textbf{Comparisons on SectLabel dataset} (Table~\ref{table:comparison:sectlabel}). Luong et at.~\cite{luong2012logical} first use Omnipage~\cite{Omnipage} to localize and recognize text lines, then predict the semantics-based label for each line. The F1 score for each class was reported. For fair comparison, we use the same set of text line bounding boxes, and use the averaged pixel-wise prediction as the label for each text line. Our model achieves better F1 scores for section heading (0.919 VS 0.916), caption (0.893 VS 0.781) and list (0.793 VS 0.712), while being capable of identifying figures and tables.

\section{Conclusion}
We proposed a multimodal fully convolutional network (MFCN) for document semantic structure extraction. The proposed model uses both visual and textual information. Moreover, we propose an efficient synthetic data generation method that yields per-pixel ground-truth. Our unsupervised auxiliary tasks help boost performance tapping into unlabeled real documents, facilitating better representation learning. We showed that both the multimodal approach and unsupervised tasks can help improve performance. Our results indicate that we have improved the state of the art on previously established benchmarks. In addition, we are publicly providing the large synthetic dataset (135,000 pages) as well as a new benchmark dataset: DSSE-200. 

\section*{Acknowledgment}
This work started during Xiao Yang's internship at Adobe Research. This work was supported by NSF grant CCF 1317560 and Adobe Systems Inc.

\clearpage
{\small
\bibliographystyle{ieee}
\bibliography{egbib}
}

\clearpage
\appendix
\section{Synthetic Document Data}\label{section:syn}
We introduced two methods to generate documents. In the first method, we generate LaTeX source files in which elements like paragraphs, figures, tables, captions, section headings and lists are randomly arranged using the ``textblock'' environment from the ``textpos'' package. Compiling these LaTeX files gives single, double, or triple-column PDFs. The generation process is summarized in Algorithm~\ref{algorithm:syn}.

\begin{algorithm}
\caption{Synthetic Document Generation}
\label{algorithm:syn}
\begin{algorithmic}[1]
\State $s$ $\leftarrow$ a string containing preamble and necessary packages of a LaTeX source file
\State Select a LaTeX source file type $T \in $ \{single-column, double-column, triple-column\}
\While{space remains on the page}
\State Select an element type $E \in $ \{figure, table, caption, section heading, list, paragraph\}
\State Select an example $e$ of type $E$
\State $s_e$ $\leftarrow$ a string of LaTeX code that generates $e$ using the ``textblock'' environment
\State $s \leftarrow s + s_e$
\EndWhile
\Ensure $s$
\Ensure A PDF document after compiling $s$
\end{algorithmic}
\end{algorithm}

\begin{figure*}
\vspace{-1.5em}
\includegraphics[width=1.0\textwidth, height=0.20\textheight]{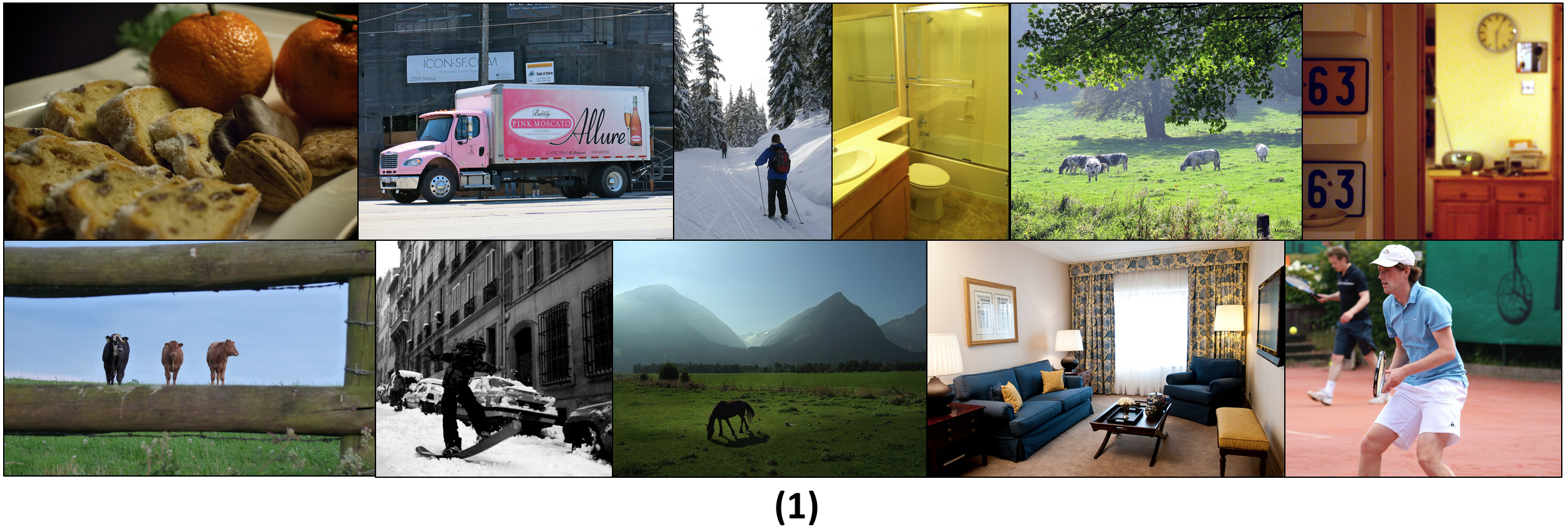}
\includegraphics[width=1.0\textwidth, height=0.20\textheight]{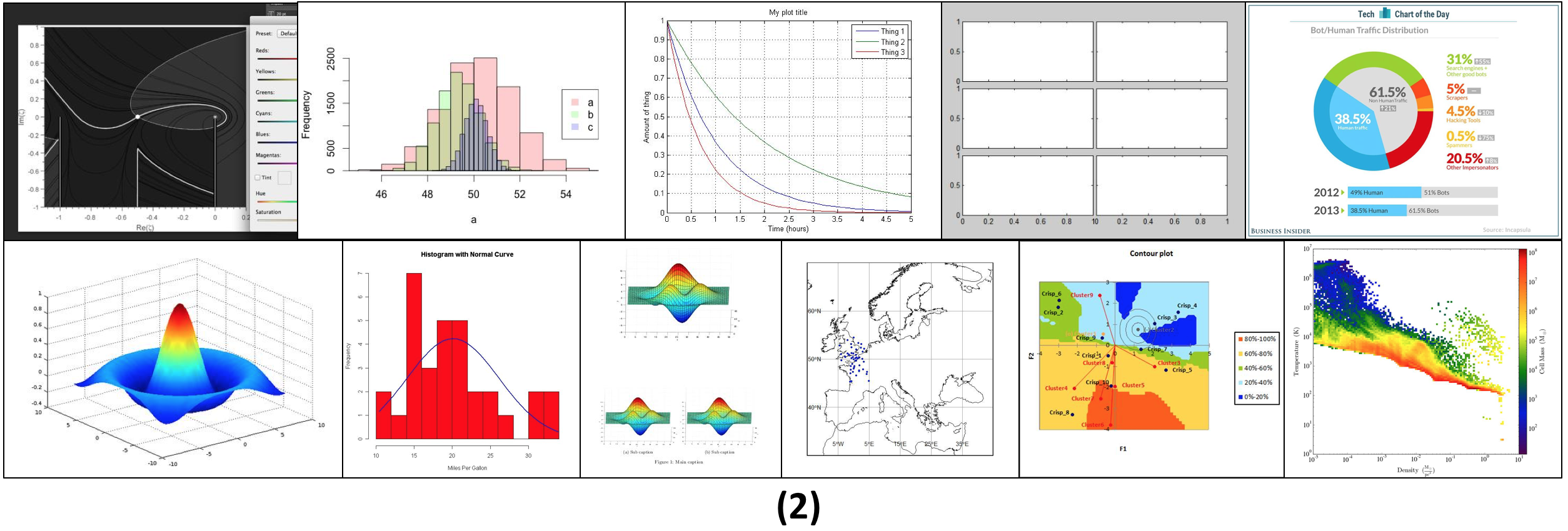}
\includegraphics[width=1.0\textwidth, height=0.20\textheight]{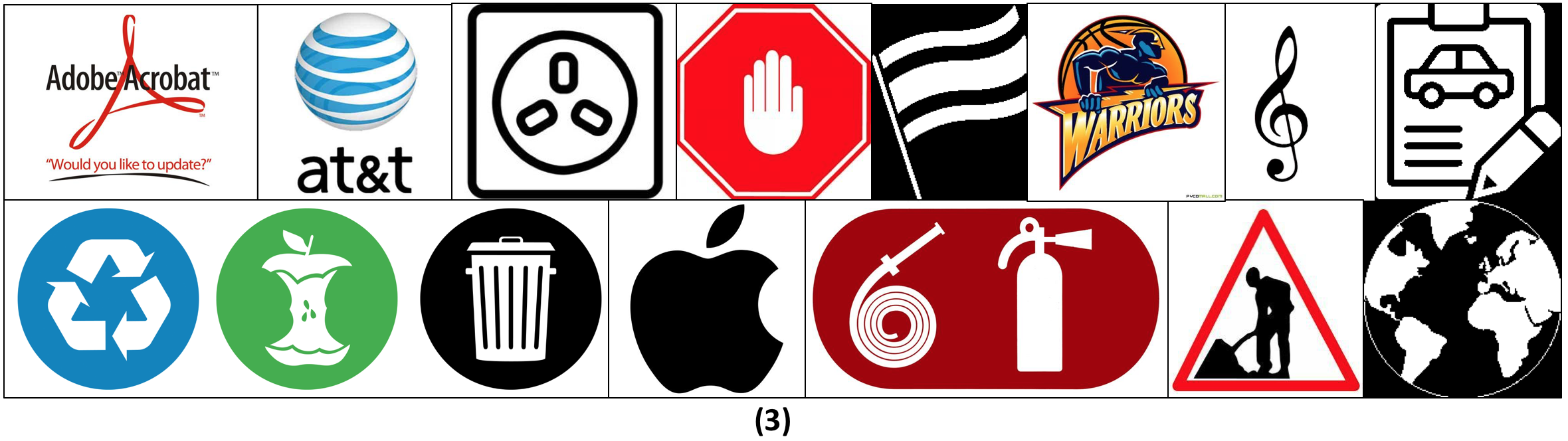}
\includegraphics[width=1.0\textwidth, height=0.35\textheight]{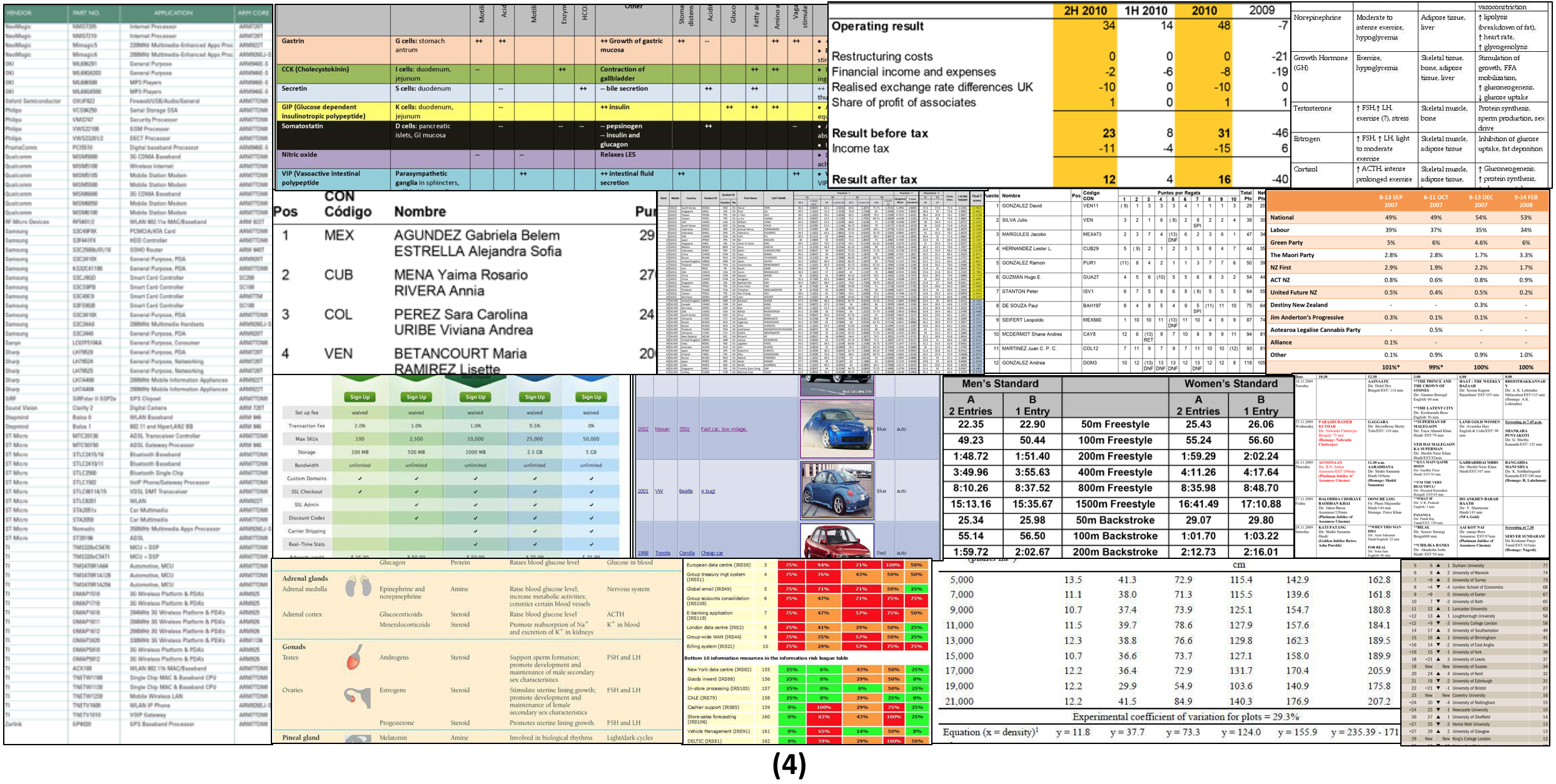}
\vspace{-1.5em}
\caption{Sample figures and tables used in synthetic documents generation. (1) Natural images from MS COCO dataset. (2) Academic-style figures from web image search. (3) Symbols and graphic drawings from web image search. (4) Tables from web image search.}
\label{figure:supplementary:elements}
\end{figure*}

Elements in a document are carefully selected following the guidelines below. Figure~\ref{figure:supplementary:elements} shows several examples of the figures and tables used in the synthetic data generation.
\begin{itemize}
	\item Candidate figures include natural images from MS COCO~\cite{lin2014microsoft}, academic-style figures and graphic drawings downloaded using web image search. 
	\item Candidates tables include table images downloaded using web image search. Various queries are used to increase the diversity of downloaded tables.
	\item For paragraphs, we randomly sample sentences from a 2016 English Wikipedia dump~\cite{Wikipedia}.
	\item For section headings, we sample sentences and phrases that are section or subsection headings in the ``Contents'' block in a Wikipedia page.
	\item For lists, we sample list items from Wikipedia pages, ensuring that all items in a list come from the same Wikipedia page.
	\item For captions, we either use the associated caption (for images from MS COCO) or the title of the image in web image search, which can be found in the span with class name ``irc\_pt''.
\end{itemize}

\begin{figure}
\vspace{-1em}
\includegraphics[width=1.\linewidth]{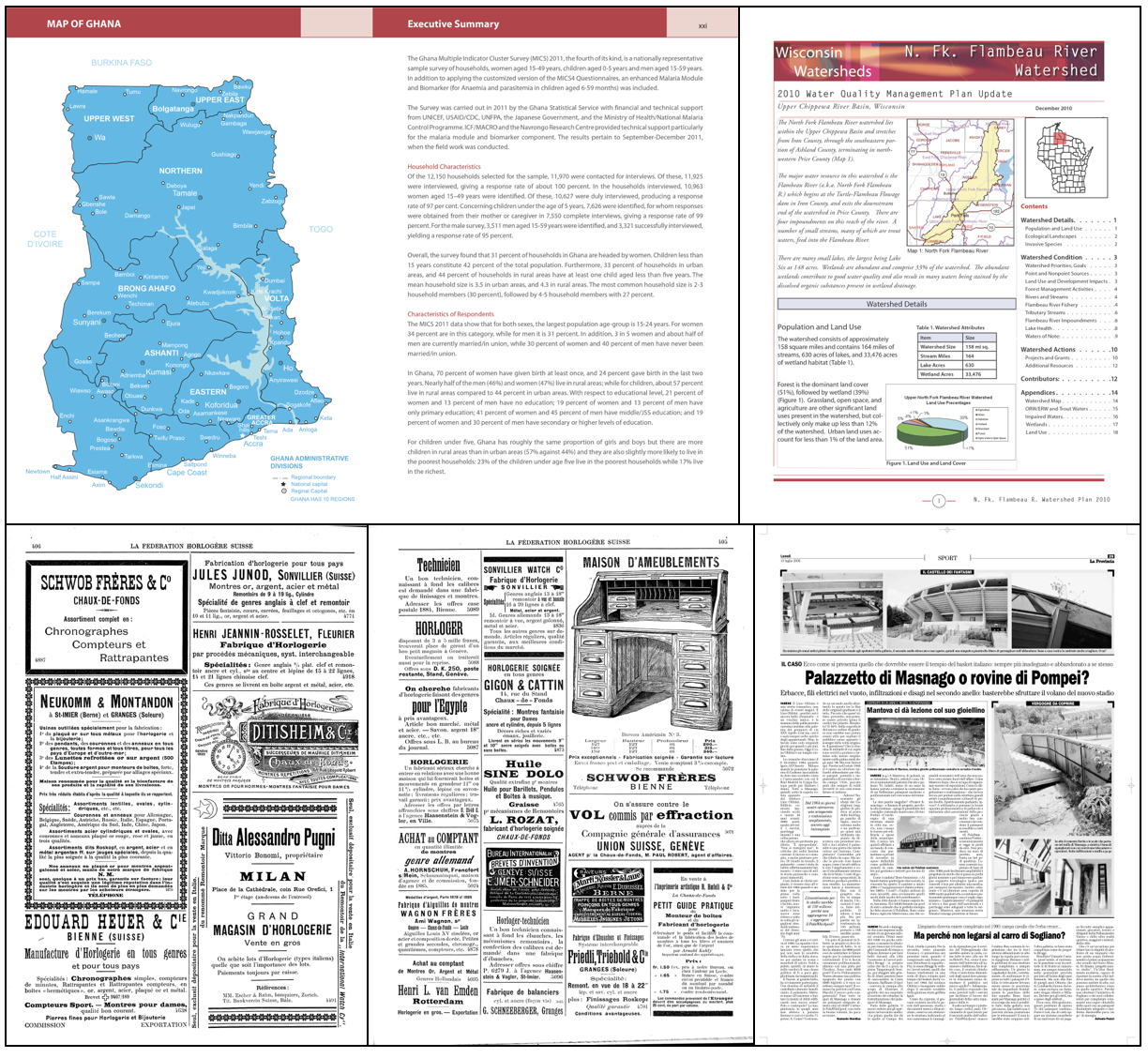}
\caption{Examples of documents with complicated layout. We labeled regions in each document and then randomly replaced them with a standalone paragraph, figure, table, caption, section heading or list, as described in Sec.~\ref{section:syn}}
\label{figure:supplementary:hard}
\vspace{-1em}
\end{figure}

In the second document generation method, we collected and labeled 271 documents with varied, complicated layouts. We then randomly replaced each element with a standalone paragraph, figure, table, caption, section heading or list generated as stated above. Figure~\ref{figure:supplementary:hard} shows several examples from the 271 documents.

\section{Visualizing the Segmentation Results}
Each pixel $p$ in the model's output layer is assigned the color of the most likely class label $l$. The RGB value of that color is then weighted by the probability $\text{P}(l)$. 

\section{Post-processing}
We apply an optional post-processing step to clean up segment masks for documents in PDF format. First, we obtain candidate bounding boxes by using the auto-tagging capabilities of Adobe Acrobat \cite{Acrobat} and parsing the results. Boxes are stored in a tree structure, and each node's  box can be a TextRun (a sequence of characters), TextLine (potentially a text line), Paragraph (potentially a paragraph) or Container (potentially figures or tables). Note that we ignore the semantic meanings associated with these boxes and only use the boxes as candidate bounding boxes in post-processing. Figure~\ref{figure:supplementary:synthetic1} (2) and \ref{figure:supplementary:synthetic2} (2) illustrate candidate bounding boxes for each document.

\begin{algorithm}
\caption{Segmentation Post-processing}
\label{algorithm:post}
\begin{algorithmic}[1]
\Require $P$ $\leftarrow$ probability map, $P(u, v) \in R^{|C|}$ is a vector containing the probability of each class $c \in C$ at location $(u, v)$
\Require $Boxes$ $\leftarrow$ candidate bounding boxes
\State $S$ $\leftarrow$ segmentation to be generated
\For{each location $(x, y) \in S$}
\State $S(x, y)$ $\leftarrow$ background
\EndFor
\For{each $b \in Boxes$} \Comment{parent box comes before child boxes}
\State $\bar{p} \leftarrow \sum_{(u, v) \in b} P(u, v)$
\State $l \leftarrow \text{argmax} \ \bar{p}$
\For{each location $(u, v) \in b$}
\If{S(u, v) is background}
\State $S(u, v) \leftarrow l$
\EndIf
\EndFor
\EndFor
\Ensure $S$
\end{algorithmic}
\end{algorithm}

Using these bounding box candidates, we refine the segmentation masks by first calculating the average class probability for pixels belonging to the same box, followed by assigning the most likely label to these pixels. The process is summarized in Algorithm~\ref{algorithm:post}.

\section{Additional Visualization Results}
\begin{figure*}
\vspace{-1em}
\includegraphics[width=1.0\textwidth]{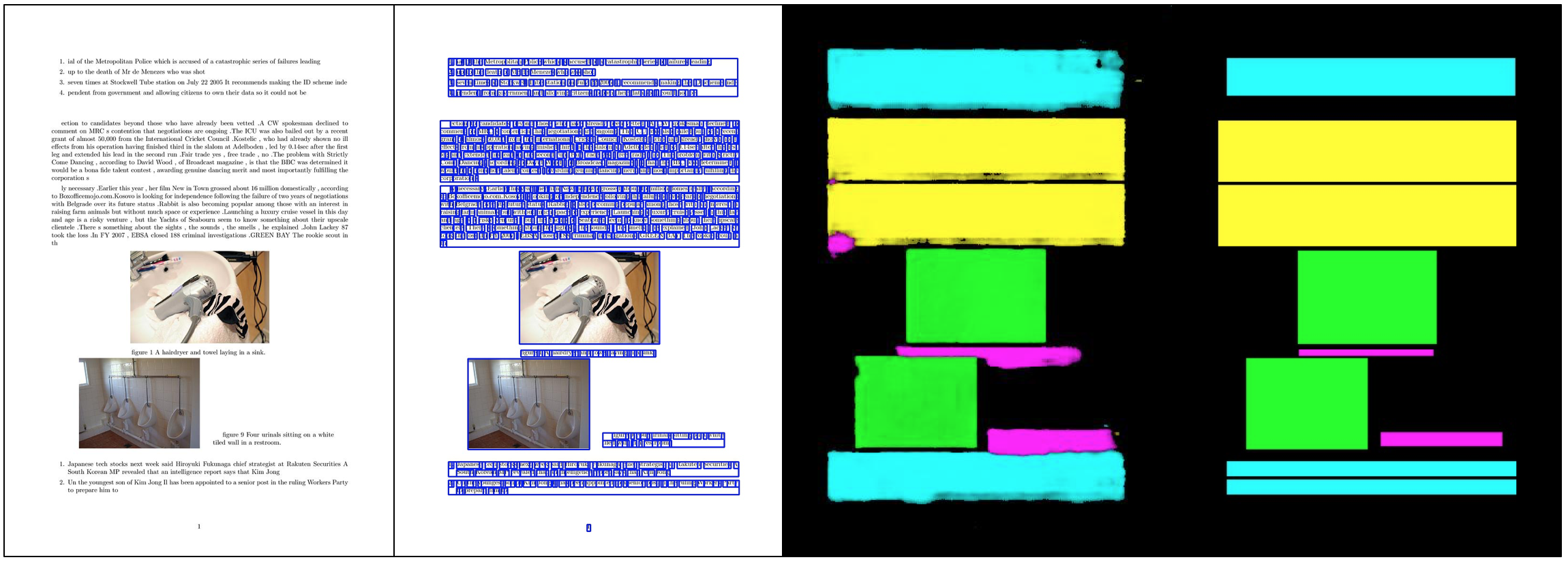}
\includegraphics[width=1.0\textwidth]{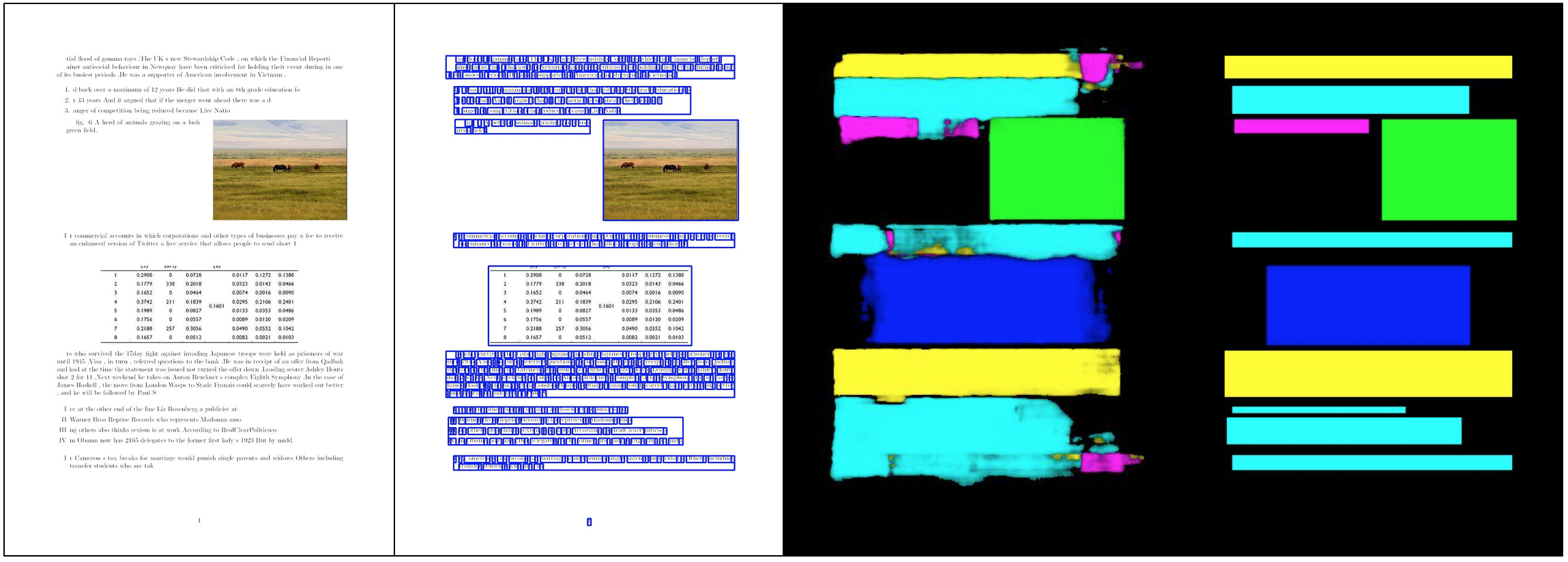}
\includegraphics[width=1.0\textwidth]{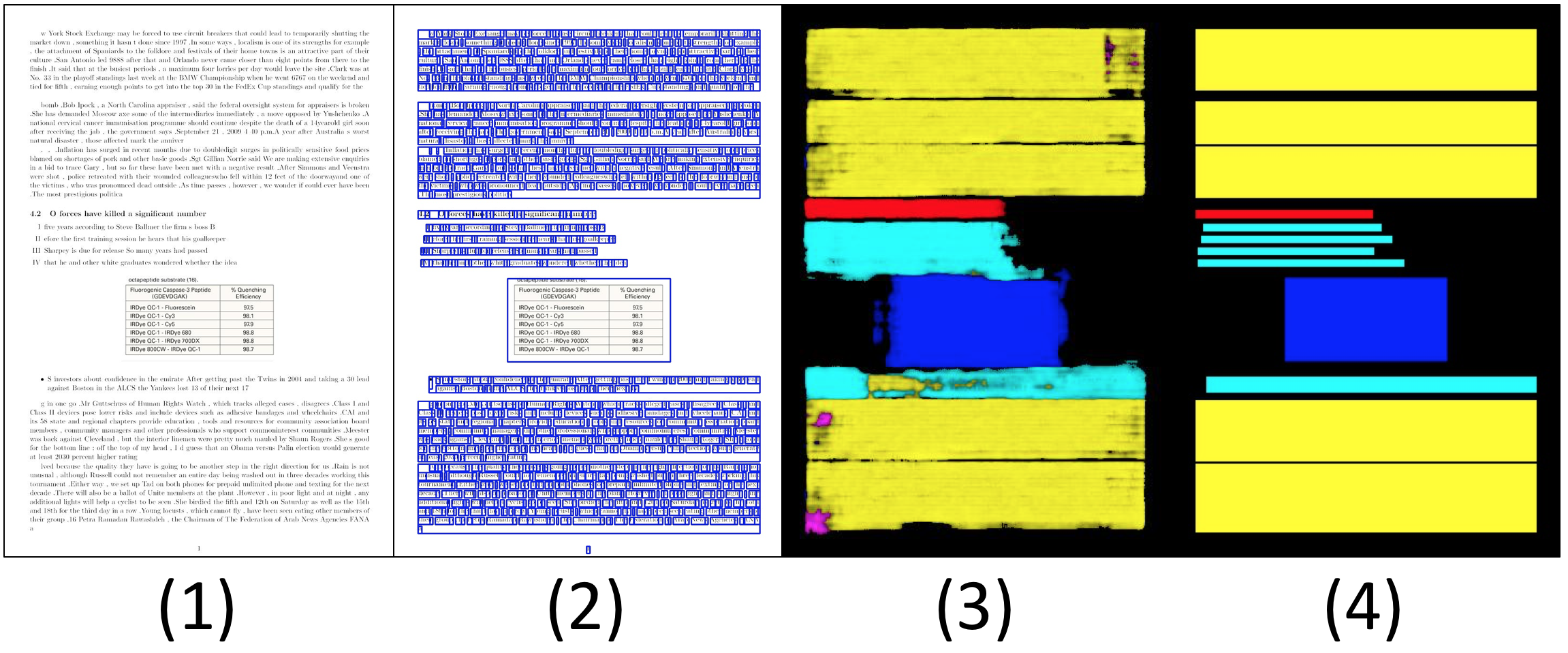}
\vspace{-1em}
\caption{Synthetic documents and the corresponding segmentations. (1) Input synthetic documents. (2) Candidate bounding boxes obtained by parsing the PDF rendering commands. (3) Raw segmentation outputs. (4) Segmentations after post-processing. Segmentation label colors are: \colorbox{green}{\strut \bf figure}, \colorbox{blue}{\strut \textcolor{white}{\bf table}}, \colorbox{red}{\strut \textcolor{white}{\bf section heading}}, \colorbox{mymagenta}{\strut \textcolor{white}{\bf caption}}, \colorbox{mycyan}{\strut \bf list} and \colorbox{yellow}{\strut \bf paragraph}.}
\label{figure:supplementary:synthetic1}
\end{figure*}

\begin{figure*}
\vspace{-1em}
\includegraphics[width=1.0\textwidth]{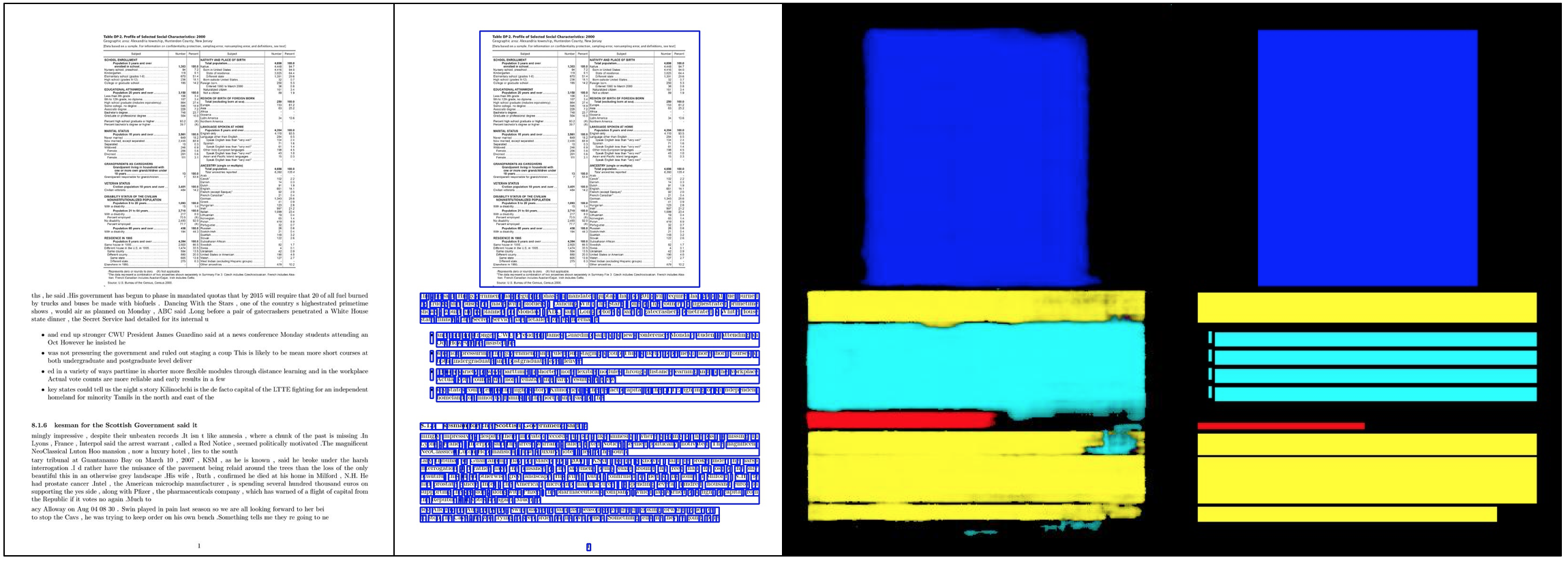}
\includegraphics[width=1.0\textwidth]{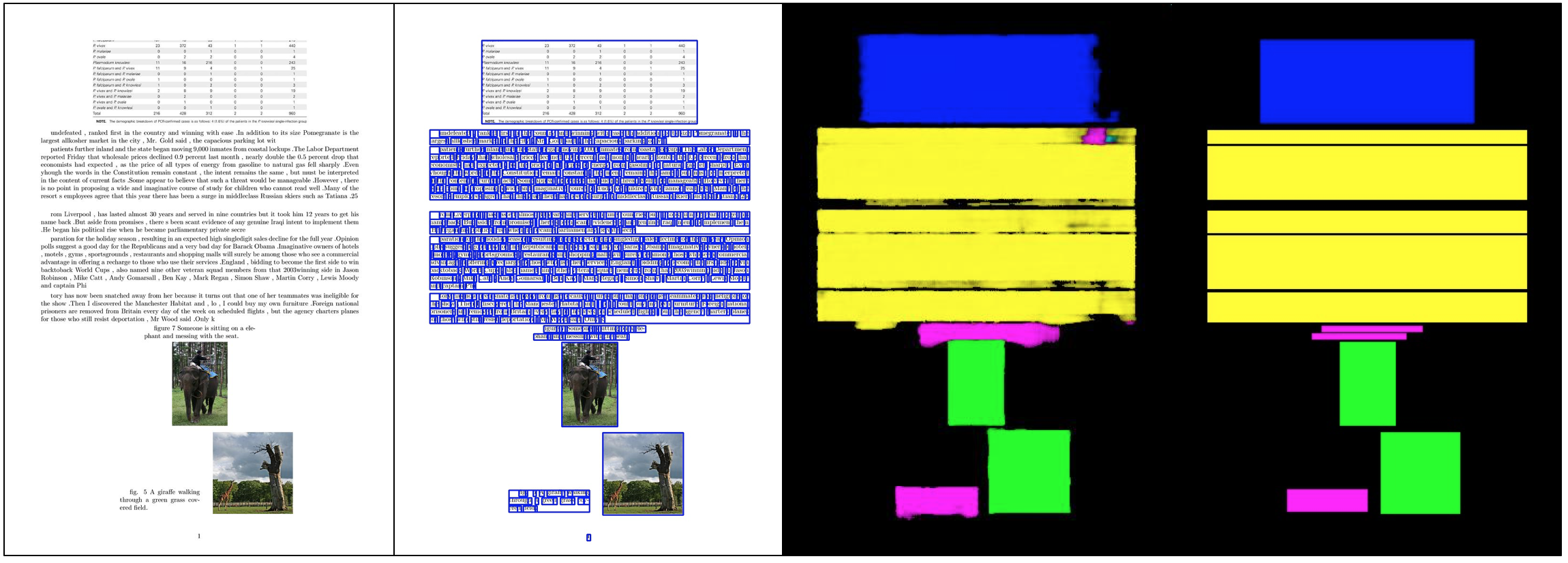}
\includegraphics[width=1.0\textwidth]{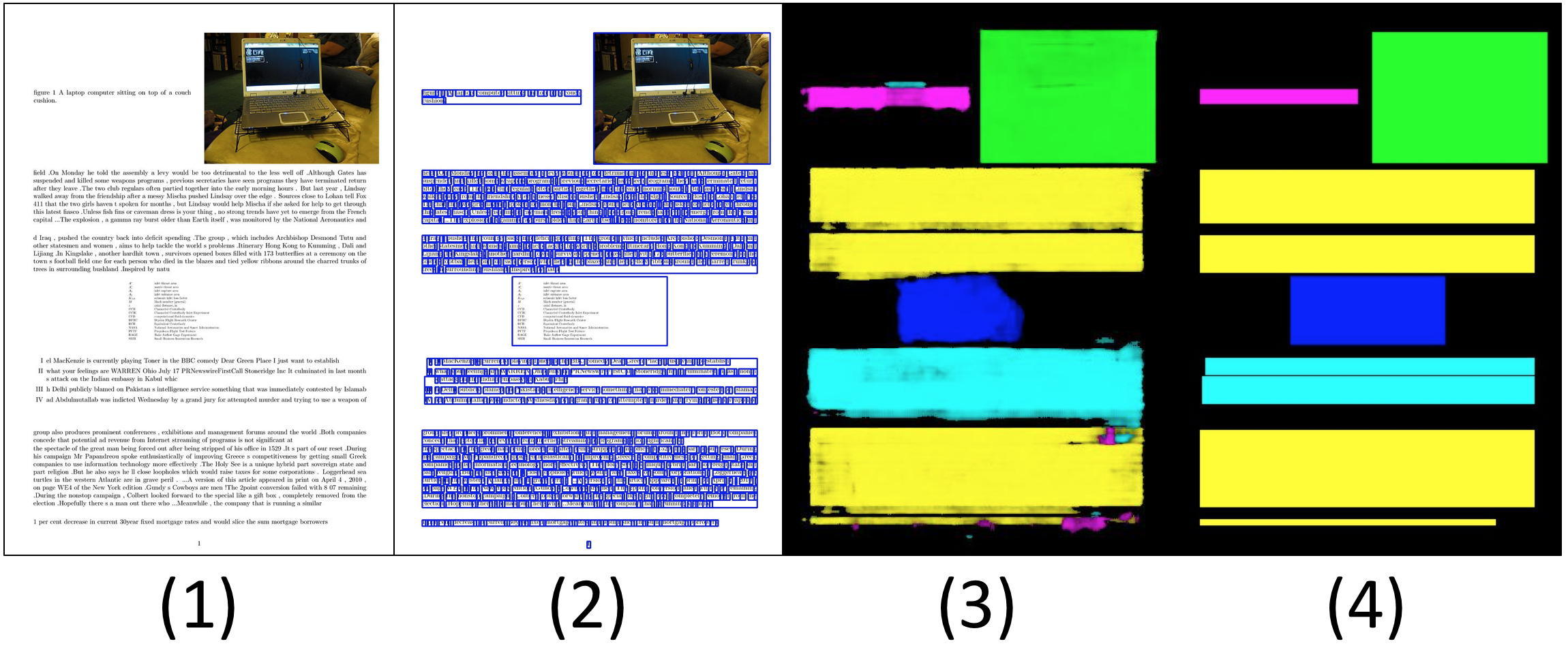}
\vspace{-1em}
\caption{Synthetic documents and the corresponding segmentations. (1) Input synthetic documents. (2) Candidate bounding boxes obtained by parsing the PDF rendering commands. (3) Raw segmentation outputs. (4) Segmentations after post-processing. Segmentation label colors are: \colorbox{green}{\strut \bf figure}, \colorbox{blue}{\strut \textcolor{white}{\bf table}}, \colorbox{red}{\strut \textcolor{white}{\bf section heading}}, \colorbox{mymagenta}{\strut \textcolor{white}{\bf caption}}, \colorbox{mycyan}{\strut \bf list} and \colorbox{yellow}{\strut \bf paragraph}.}
\label{figure:supplementary:synthetic2}
\end{figure*}

\begin{figure*}
\vspace{-1em}
\includegraphics[width=1.0\textwidth, height=0.23\textheight]{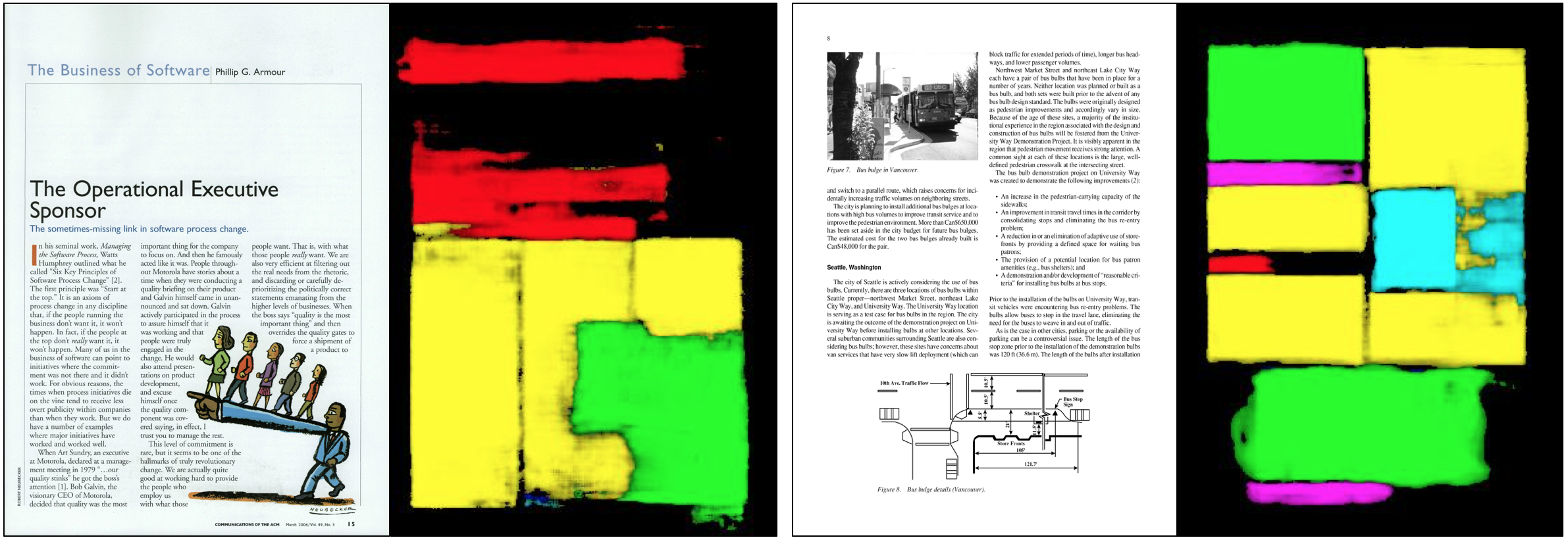}
\includegraphics[width=1.0\textwidth, height=0.23\textheight]{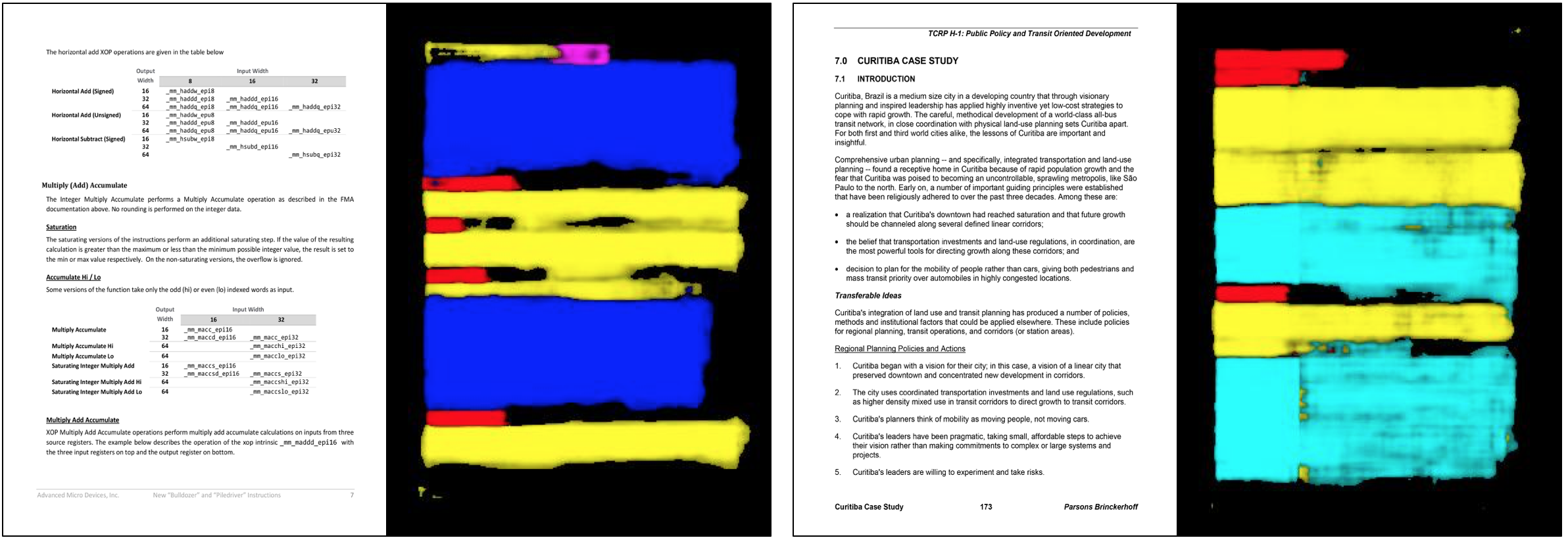}
\includegraphics[width=1.0\textwidth, height=0.23\textheight]{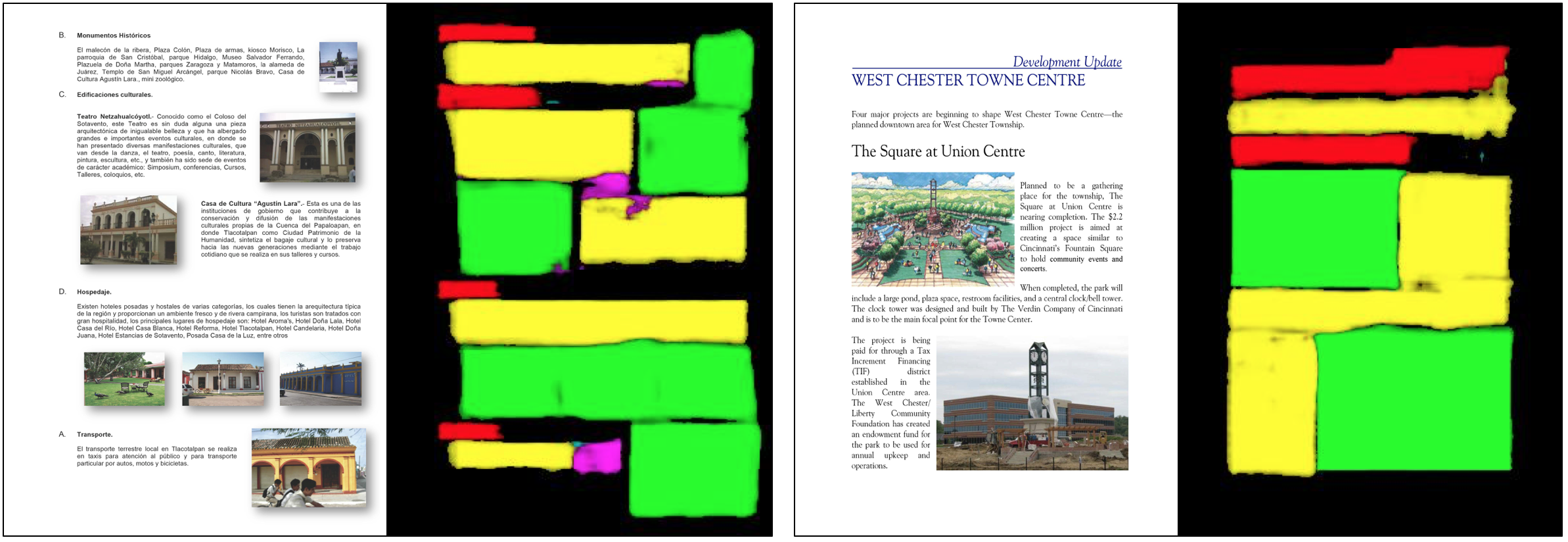}
\includegraphics[width=1.0\textwidth, height=0.23\textheight]{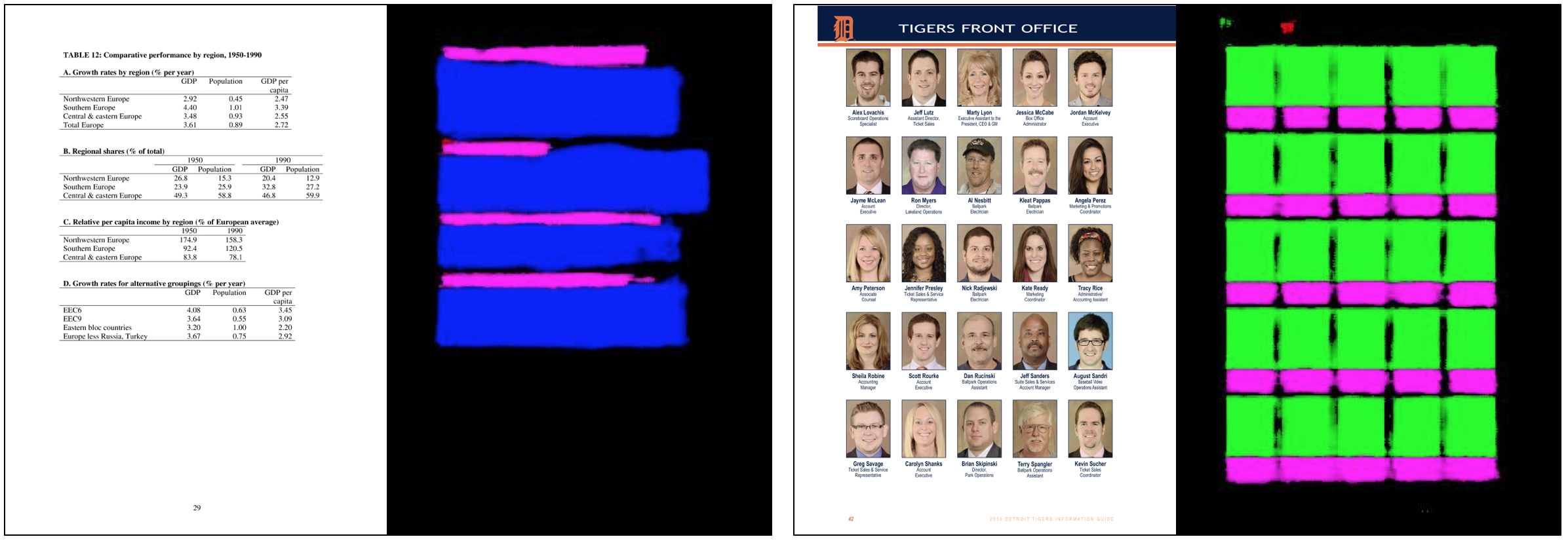}
\vspace{-1em}
\caption{Real documents and the corresponding segmentations. Segmentation label colors are: \colorbox{green}{\strut \bf figure}, \colorbox{blue}{\strut \textcolor{white}{\bf table}}, \colorbox{red}{\strut \textcolor{white}{\bf section heading}}, \colorbox{mymagenta}{\strut \textcolor{white}{\bf caption}}, \colorbox{mycyan}{\strut \bf list} and \colorbox{yellow}{\strut \bf paragraph}.}
\label{figure:supplementary:real}
\end{figure*}

Figures~\ref{figure:supplementary:synthetic1} and~\ref{figure:supplementary:synthetic2} show additional visualization examples of synthetic documents, and Figure~\ref{figure:supplementary:real} shows additional examples of real documents.

\end{document}